\documentclass[journal]{IEEEtran}

\usepackage{amsmath,amsfonts,bm}

\def\1{\bm{1}}

\def\vf{{\bm{f}}}

\def\vh{{\bm{h}}}

\def\vx{{\bm{x}}}

\DeclareMathAlphabet{\mathsfit}{\encodingdefault}{\sfdefault}{m}{sl}
\SetMathAlphabet{\mathsfit}{bold}{\encodingdefault}{\sfdefault}{bx}{n}
\newcommand{\tens}[1]{\bm{\mathcal{#1}}}

\def\tI{{\tens{I}}}

\def\tX{{\tens{X}}}
\def\tY{{\tens{Y}}}
\def\tZ{{\tens{Z}}}

\usepackage{amsmath,amssymb}
\usepackage{graphicx}
\usepackage{algorithm,algorithmic}
\usepackage{color}
\usepackage{url}
\usepackage[pagebackref=true,breaklinks=true,letterpaper=true,colorlinks,bookmarks=false]{hyperref}

\newcommand{\bbR}[1]{\mathbb{R}^{#1}}
\newcommand{\ie}{{\em i.e. }}
\newcommand{\eg}{{\em e.g. }}

\begin{document}

\title{Manifold Modeling in Embedded Space: A Perspective for Interpreting Deep Image Prior}

\author{Tatsuya Yokota,
  Hidekata Hontani,
  Qibin Zhao,
  and Andrzej Cichocki,
\thanks{Tatsuya Yokota and Hidekata Hontani are with Nagoya Institute of Technology, Japan. e-mail: t.yokota@nitech.ac.jp.}
\thanks{Qibin Zhao is with RIKEN Center for Advanced Intelligence Project, Japan.}
\thanks{Andrzej Cichocki is with Skolkovo Institute of Science and Technology, Russia and RIKEN Center for Advanced Intelligence Project, Japan.}
}

\markboth{Yokota \MakeLowercase{\textit{et al.}}: Manifold Modeling in Embedded Space: A Perspective for Interpreting Deep Image Prior}
{Yokota \MakeLowercase{\textit{et al.}}: Manifold Modeling in Embedded Space: A Perspective for Interpreting Deep Image Prior}

\IEEEcompsoctitleabstractindextext{
  \begin{abstract}
Deep image prior (DIP) \cite{ulyanov2018deep}, which utilizes a deep convolutional network (ConvNet) structure itself as an image prior, has attracted attentions in computer vision and machine learning communities.  It empirically shows the effectiveness of ConvNet structure for various image restoration applications.  However, why the DIP works so well is still unknown, and why convolution operation is useful for image reconstruction or enhancement is not very clear. In this study, we tackle these questions. The proposed approach is dividing the convolution into ``delay-embedding'' and ``transformation (\ie encoder-decoder)'', and proposing a simple, but essential, image/tensor modeling method which is closely related to dynamical systems and self-similarity. The proposed method named as manifold modeling in embedded space (MMES) is implemented by using a novel denoising-auto-encoder in combination with multi-way delay-embedding transform. In spite of its simplicity, the image/tensor completion, super-resolution, deconvolution, and denoising results of MMES are quite similar even competitive to DIP in our extensive experiments, and these results would help us for reinterpreting/characterizing the DIP from a perspective of ``low-dimensional patch-manifold prior''.
\end{abstract}
}

\maketitle

\IEEEdisplaynotcompsoctitleabstractindextext

\IEEEpeerreviewmaketitle

\section{Introduction}\label{sec:intro}
The most important piece of information for image/tensor restoration would be the ``prior'' which usually converts the optimization problems from ill-posed to well-posed, and/or gives some robustness for specific noises and outliers. Many priors were studied in computer science problems such as low-rank representation \cite{pearson1901liii,hotelling1933analysis,hitchcock1927expression,tucker1966some}, smoothness \cite{grimson1981images,poggio1985computational,li1994markov}, sparseness \cite{tibshirani1996regression}, non-negativity \cite{lee1999learning,cichocki2009nonnegative}, statistical independence \cite{hyvarinen2004independent}, and so on. Particularly in today's computer vision and machine learning problems, total variation (TV) \cite{guichard1998total,vogel1998fast}, low-rank representation \cite{liu2013tensor,ji2010robust,zhao2015bayesian,hu2016twist,wang2017efficient,wu2018fused,zhang2018nonconvex,shi2018feature,xue2019enhanced,zhang2019accurate}, and non-local similarity \cite{buades2005non,dabov2007image} priors are often used for image modeling.
These priors can be obtained by analyzing basic properties of natural images, and categorized as ``unsupervised image modeling''.

\begin{figure}[t]
\centering
\includegraphics[width=0.48\textwidth]{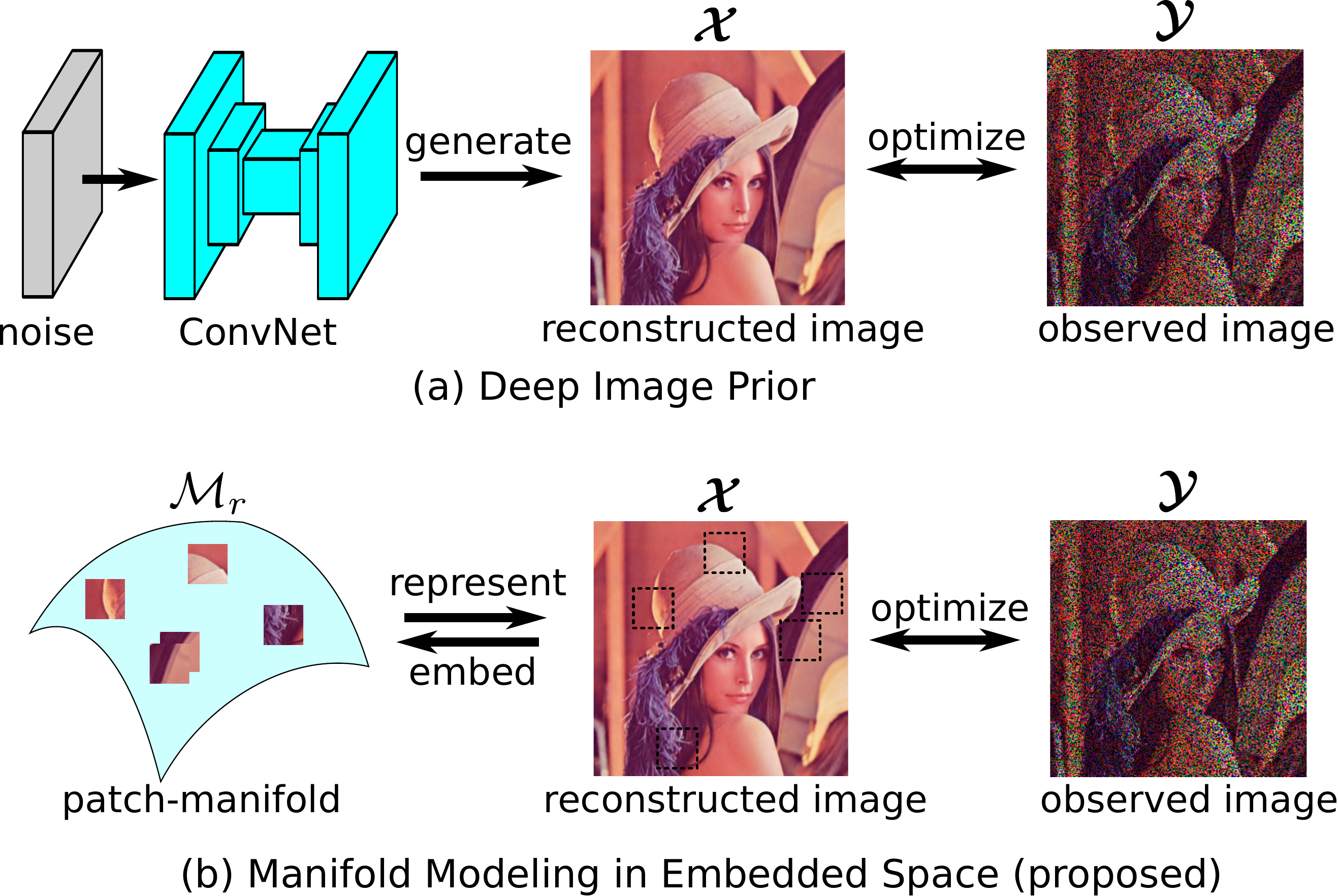}
\caption{Conceptual illustration of deep image prior and the proposed manifold modeling in embedded space. A case for image inpainting task.}\label{fig:concept}
\end{figure}

By contrast, the deep image prior (DIP) \cite{ulyanov2018deep} has been come from a part of ``supervised'' or ``data-driven'' image modeling framework (\ie deep learning) although the DIP itself is one of the state-of-the-art unsupervised image restoration methods.
Fig.~\ref{fig:concept}(a) shows a conceptual illustration of the method of DIP in image inpainting task.
The method of DIP can be simply explained to only optimize an {\em untrained} (\ie randomly initialized) fully convolutional generator network (ConvNet) for minimizing squares loss between its generated image and an observed image (\eg noisy/incomplete image), and stop the optimization before the overfitting.  In \cite{ulyanov2018deep}, the authors explained the reason why a high-capacity ConvNet can be used as a prior by the following statement: {\em Network resists ``bad'' solutions and descends much more quickly towards naturally-looking images}, and its phenomenon of ``impedance of ConvNet'' was confirmed by toy experiments.
However, most researchers could not be fully convinced from only above explanation because it is just a part of whole.
One of the essential questions is why is it ConvNet? or in  more practical perspective, to explain what is ``priors in DIP'' with simple and clear words (like smoothness, sparseness, low-rank etc) is very important.

\begin{figure}[t]
\centering
\includegraphics[width=0.48\textwidth]{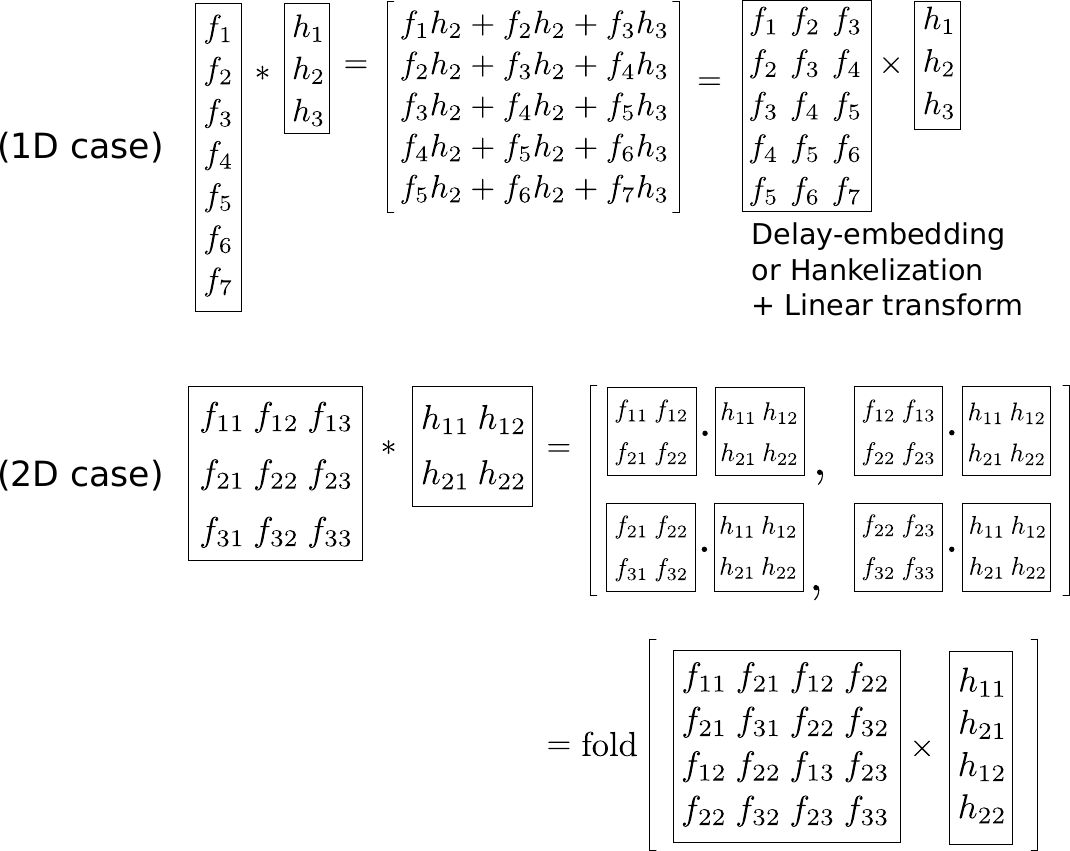}
\caption{Decomposition of 1D and 2D convolutions: Valid convolution can be divided into delay-embedding/Hankelization and linear transformation. 1D valid convolution of $\vf$ with kernel $\vh = [h_1, h_2, h_3]$ can be provided by matrix-vector product of the Hankel matrix and $\vh$.
In similar way, 2D valid convolution can be provided by matrix-vector product of the block Hankel matrix and unfolded kernel.}\label{fig:conv_decomp}
\end{figure}

In this study, we tackle the question why ConvNet is essential as an image prior, and try to translate the ``deep image prior'' with words.
For this purpose, we divide the convolution operation into ``embedding'' and ``transformation'' (see Fig.~\ref{fig:conv_decomp}).
Here, the ``embedding'' stands for delay/shift-embedding (\ie Hankelization) which is a copy/duplication operation of image-patches by sliding window of patch size $(\tau,\tau)$.
The embedding/Hankelization is a preprocessing to capture the delay/shift-invariant feature (\eg non-local similarity) of signals/images.
This ``transformation'' is basically linear transformation in a simple convolution operation, and it also indicates some non-linear transformation from the artificial neural network perspective.

\begin{figure}[t]
\centering
\includegraphics[width=0.47\textwidth]{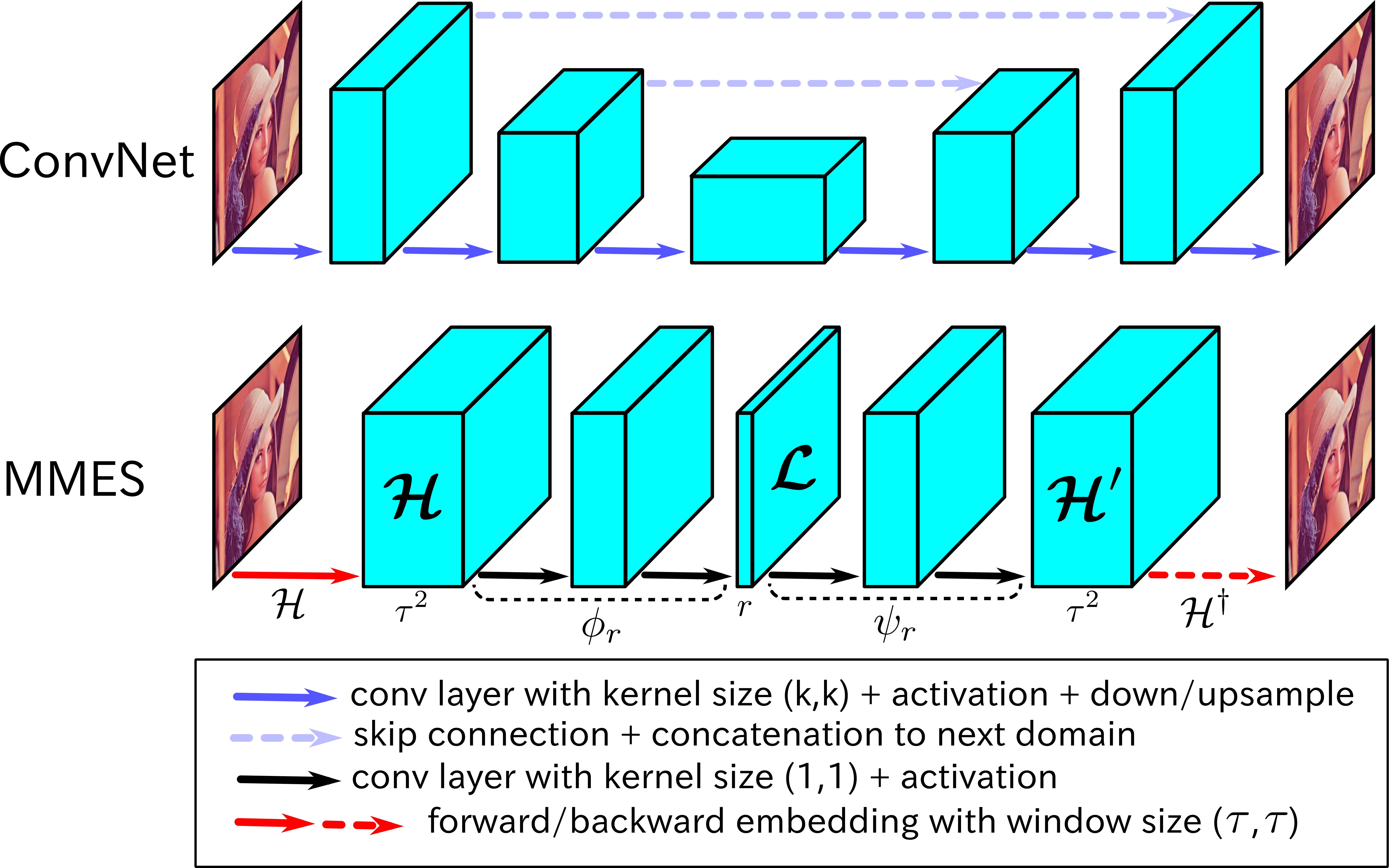}
\caption{Comparison of typical auto-encoder ConvNet and the proposed MMES network.}\label{fig:convnet}
\end{figure}

To simplify the complicated ``encoder-decoder'' structure of ConvNet used in DIP, we consider the following network structure: Embedding $\mathcal{H}$ (linear), encoding $\phi_r$ (non-linear), decoding $\psi_r$ (non-linear), and backward embedding $\mathcal{H}^\dagger$ (linear) (see Fig.~\ref{fig:convnet}).
Note that its encoder-decoder part $(\phi_r,\psi_r)$ is just a simple multi-layer perceptron along the filter domain (\ie manifold learning), and it is sandwitched between forward and backward embedding ($\mathcal{H}$, $\mathcal{H}^\dagger$).
Hence, the proposed network can be characterized by Manifold Modeling in Embedded Space (MMES).
The proposed MMES is designed as simple as possible while keeping a essential ConvNet structure.
Some parameters $\tau$ and $r$ in MMES are corresponded with a kernel size and a filter size in ConvNet.

Fig.~\ref{fig:convnet} shows network structures of ConvNet and MMES.
When we set the horizontal dimension of hidden tensor $\bm{\mathcal{L}}$ with $r$, each $\tau^2$-dimensional fiber in $\bm{\mathcal{H}}$, which is a vectorization of each $(\tau,\tau)$-patch of an input image, is encoded into $r$-dimensional space.
Note that the volume of hidden tensor $\bm{\mathcal{L}}$ looks to be larger than that of input/output image, but representation ability of $\bm{\mathcal{L}}$ is much lower than input/output image space since the first/last tensor ($\bm{\mathcal{H}}$,$\bm{\mathcal{H}}'$) must have Hankel structure (\ie its representation ability is equivalent to image) and the hidden tensor $\bm{\mathcal{L}}$ is reduced to lower dimensions from $\bm{\mathcal{H}}$.
Here, we assume $r < \tau^2$, and its low-dimensionality indicates the existence of similar ($\tau,\tau$)-patches (\ie self-similarity) in the image, and it would provide some ``impedance'' which passes self-similar patches and resist/ignore others.
Each fiber of Hidden tensor $\bm{\mathcal{L}}$ represents a coordinate on the patch-manifold of image.

It should be noted that the MMES network is a special case of deep neural networks.
In fact, the proposed MMES can be considered as a new kind of auto-encoder (AE) in which convolution operations have been replaced by Hankelization in pre-processing and post-processing.
Compared with ConvNet, the forward and backward embedding operations can be implemented by convolution and transposed convolution with one-hot-filters (see Fig.~\ref{fig:mdt_conv} for details).
Note that the encoder-decoder part can be implemented by multiple convolution layers with kernel size (1,1) and non-linear activations.
In our model, we do not use convolution explicitly but just do linear transform and non-linear activation for ``filter-domain'' (\ie horizontal axis of tensors in Fig.~\ref{fig:convnet}).

The contributions in this study can be summarized as follow: (1) A new and simple approach of image/tensor modeling is proposed which translates the ConvNet, (2) effectiveness of the proposed method and similarity to the DIP are demonstrated in experiments, and (3) most importantly, there is a prospect for interpreting/characterizing the DIP as ``low-dimensional patch-manifold prior''.




\section{Related works}
Note that the idea of low-dimensional patch manifold itself has been proposed by \cite{peyre2009manifold} and \cite{osher2017low}.
Peyre had firstly formulated the patch manifold model of natural images and solve it by dictionary learning and manifold pursuit.
Osher \textit{et al}. formulated the regularization function to minimize dimension of patch manifold, and solved Laplace-Beltrami equation by point integral method.
In comparison with these studies, we decrease the dimension of patch-manifold by utilizing AE shown in Fig.~\ref{fig:convnet}.

A related technique, low-rank tensor modeling in embedded space, has been studied recently by \cite{yokota2018missing}.
However, the modeling approaches here are different: multi-linear vs non-linear manifold.
Thus, our study would be interpreted as manifold version of \cite{yokota2018missing} in a perspective of tensor completion methods.
Note that \cite{yokota2018missing} applied their model for only tensor completion task.
By contrast, we investigate here tensor completion, super-resolution, deconvolution, and denoising tasks.

Another related work is devoted to group sparse representation (GSR) \cite{zhang2014group}.  The GSR is roughly characterized as a combination of similar patch-grouping and sparse modeling which is similar to the combination of embedding and manifold-modeling.  However, the computational cost of similar patch-grouping is obviously higher than embedding, and this task is naturally included in manifold learning.

The main difference between above studies and our is the motivation: Essential and simple image modeling which can translate the ConvNet/DIP.
The proposed MMES has many connections with ConvNet/DIP such as embedding, non-linear mapping, and the training with noise.


\begin{figure*}[t]
\centering
\includegraphics[width=0.95\textwidth]{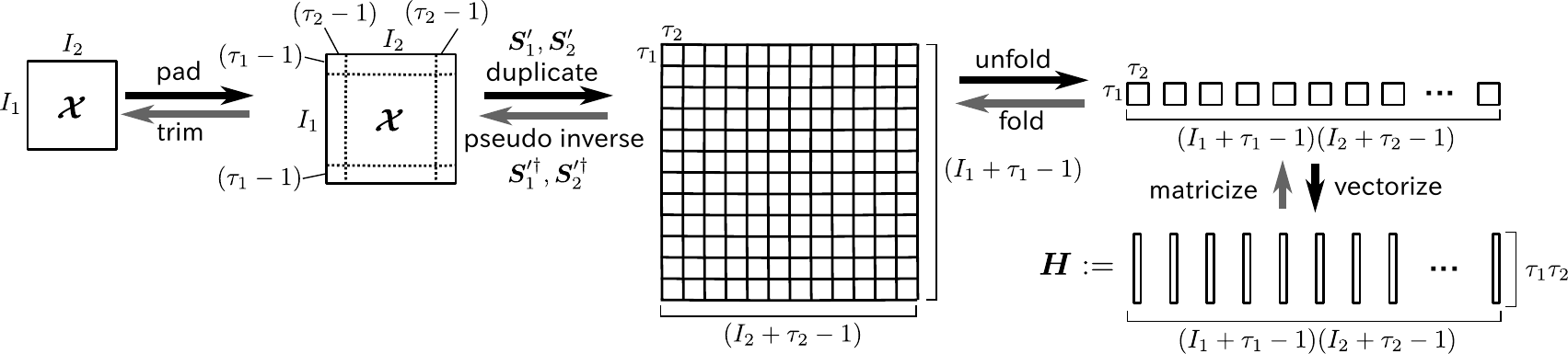}
\caption{Flow of multiway-delay-embedding operation ($N=2$).}\label{fig:mdt}
\end{figure*}

From a perspective of DIP, there are several related works.
First, the deep geometric prior \cite{williams2019deep} utilises a good properties of a multi-layer perceptron for shape reconstruction problem which efficiently learn a smooth function from 2D space to 3D space.  It helps us to understand DIP from a perspective of manifold learning.  For example, it can be used for gray scale image reconstruction if an image is regarded as point could in 3D space $(i, j, X_{ij})$.  However, this may not provide the good image reconstruction like DIP, because it just smoothly interpolates a point cloud by surface like a Volonoi interpolation.  Especially it can not provide a property of self-similarity in natural image.

Second, deep decoder \cite{heckel2018deep} reconstructs natural images from noises by non-convolutional networks which consists of linear channel/color transform, ReLU, channel/color normalization, and upsampling layers.
In contrast that DIP uses over-parameterized network, deep decoder uses under-parameterized network and shows its ability of image reconstruction.
Although deep decoder is a non-convolutional network, Authors emphasize the closed relationship between convolutional layers in DIP and upsampling layers in deep decoder.
In this literature, Authors described "If there is no upsampling layer, then there is no notion of locality in the resultant image" in deep decoder.
It implies the "locality" is the essence of image model, and the convolution/upsampling layer provides it. Furthermore, the deep decoder has a close relationship with our MMES. Note that the MMES is originally/essentially has only decoder and inverse MDT (see Eq.~\eqref{eq:generator}), and the encoder is just used for satisfying Hankel structure.  The decoder and inverse MDT in our MMES are respectively corresponding linear operation and upsampling layer in deep decoder.  Moreover, concept of under-parameterization is also similar to our MMES.

From this, we can say the essence of image model is the "locality", and its locality can be provided by "convolution", "upsampling", or "delay-embedding".
This is why the image restoration from single image with deep convolutional networks has highly attentions which are called by zero-shot learning, internal learning, or self-supervised learning \cite{shocher2018zero,lehtinen2018noise2noise,krull2019noise2void,batson2019noise2self,xu2019noisy,cha2019gan2gan,laine2019self}.

Recently, two generative models: SinGAN \cite{shaham2019singan} and InGAN \cite{shocher2019ingan} learned from only a single image, have been proposed.  Key concept of both papers is to impose the constraint for local patches of image to be natural.  From a perspective of the constraint for local patches of image, our MMES has closed relationship with these works. However, we explicitly impose a low-dimensional manifold constraint for local patches rather than adversarial training with patch discriminators.

\section{Manifold Modeling in Embedded Space}\label{sec:proposed}
Here, on the contrary to Section~\ref{sec:intro}, we start to explain the proposed method from the concept of MMES, and we systematically derive the MMES structure from it.
Conceptually, the proposed tensor reconstruction method can be formulated by
\begin{align}
  \mathop{\text{minimize}}_{{\tX}} & \  || {\tY} - {\mathcal F}({\tX}) ||_F^2, \notag \\
                    \text{ s.t. } & \  {\mathcal H}({\tX}) = [\bm h_1, \bm h_2, ..., \bm h_T] =: \bm H, \label{eq:concept}\\
                                & \ \bm h_t \in \mathcal{M}_r \text{ for } t=1,2,...,T, \notag
\end{align}
where ${\tY} \in \bbR{J_1 \times J_2 \times \cdots \times J_N}$ is an observed corrupted tensor, ${\tX} \in \bbR{I_1 \times I_2 \times \cdots \times I_N}$ is an estimated tensor, ${\mathcal F}: \bbR{I_1 \times I_2 \times \cdots \times I_N} \rightarrow  \bbR{J_1 \times J_2 \times \cdots \times J_N}$ is a linear operator which represents the observation system, ${\mathcal H}: \bbR{I_1 \times I_2 \times \cdots \times I_N} \rightarrow \bbR{D \times T}$ is padding and Hankelization operator with sliding window of size $(\tau_1, \tau_2, ..., \tau_N)$, and we impose each column of matrix ${\bm H}$ can be sampled from an $r$-dimensional manifold ${\mathcal M}_r$ embedded in $D$-dimensional Euclidean space (see Appendix B for details).  We have $r \leq D$.
For simplicity, we putted $ D := \prod_n \tau_n$ and $T:=\prod_n (I_n+\tau_n-1)$.
For tensor completion task, ${\mathcal F} := P_\Omega$ is a projection operator onto support set $\Omega$ so that the missing elements are set to be zero.
For super-resolution task, ${\mathcal F}$ is a down-sampling operator of images/tensors.
For deconvolution task, ${\mathcal F}$ is a convolution operator with some blur kernels.
For denoising task, ${\mathcal F}$ is an identity map.
Fig.~\ref{fig:concept}(b) shows the concept of proposed manifold modeling in case of image inpainting/completion (\ie $N=2$).
We minimize the distance between observation ${\tY}$ and reconstruction ${\tX}$ with its support $\Omega$, and all patches in ${\tX}$ should be included in some restricted manifold $\mathcal{M}_r$.
In other words, ${\tX}$ is represented by the patch-manifold, and the property of the patch-manifold can be image priors.  For example, low dimensionality of patch-manifold restricts the non-local similarity of images/tensors, and it would be related with ``impedance'' in DIP.  We model ${\tX}$ indirectly by designing the properties of patch-manifold $\mathcal{M}_r$.

\begin{figure}[t]
\centering
\includegraphics[width=0.35\textwidth]{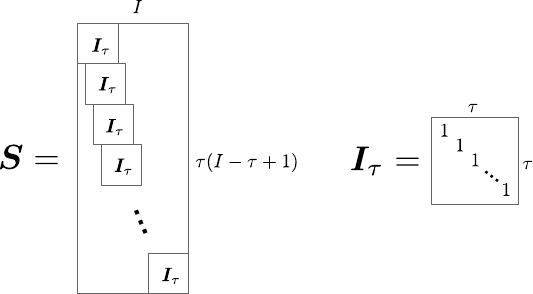}
\caption{Duplication matrix.  In case that we have $I$ columns, it consists of $(I-\tau+1)$ identity matrices of size $(\tau,\tau)$.}\label{fig:dup}
\end{figure}

\subsection{Multiway-delay embedding for tensors}\label{sec:MDT}
Multiway-delay embedding transform (MDT) is a multi-way generalization of Hankelization proposed by \cite{yokota2018missing}.

In \cite{yokota2018missing}, MDT is defined by using the multi-linear tensor product with multiple duplication matrices and tensor reshaping.
Basically, we use the same operation, but a padding operation is added.
Thus, the multiway-delay embedding used in this study is defined by
\begin{align}
  {\mathcal H}({\tX}) := \text{unfold}_{(D,T)}(\text{pad}_{\bm \tau}({\tX}) \times_1 \bm S_1 \cdots \times_N \bm S_N),
\end{align}
where $\text{pad}_{\bm \tau} : \bbR{I_1 \times \cdots \times I_N} \rightarrow  \bbR{(I_1+2(\tau_1-1)) \times \cdots \times (I_N+2(\tau_N-1))}$ is a $N$-dimensional reflection padding operator of tensors, $\bm S_n \in \bbR{\tau_n(I_n + \tau_n - 1) \times (I_n + 2(\tau_n -1))}$ is a duplication matrix (see Fig.~\ref{fig:dup}), $\text{unfold}_{(D, T)} : \bbR{\tau_1(I_1 +\tau_1 - 1) \times \cdots \times \tau_N(I_N +\tau_N - 1)} \rightarrow \bbR{D \times T}$ is an unfolding operator which outputs a matrix from an input $N$-th order tensor.

For example, our Hankelization with reflection padding of $\vx = [x_1, x_2, ..., x_7]^T$ with $\tau=3$ is given by
\begin{align}
  & [x_1, x_2, x_3, x_4, x_5, x_6, x_7]^T \notag \\
  & \mathop{\longrightarrow}^{\text{pad}_3} [x_3, x_2, x_1, x_2, x_3, x_4, x_5, x_6, x_7, x_6, x_5]^T \notag \\
  & \mathop{\longrightarrow}^{\text{Hankelize}}
  \begin{pmatrix}
    x_3 & x_2 & x_1 & x_2 & x_3 & x_4 & x_5 & x_6 & x_7 \\
    x_2 & x_1 & x_2 & x_3 & x_4 & x_5 & x_6 & x_7 & x_6 \\
    x_1 & x_2 & x_3 & x_4 & x_5 & x_6 & x_7 & x_6 & x_5
  \end{pmatrix}.
\end{align}

Fig.~\ref{fig:mdt} shows an example of our multiway-delay embedding in case of second order tensors.  The overlapped patch grid is constructed by multi-linear tensor product with $\bm S_n$.  Finally, all patches are splitted, lined up, and vectorized.


The Moore-Penrose pseudo inverse of $\mathcal{H}$ is given by
\begin{align}
  {\mathcal H}^{\dagger}(\bm H) = \text{trim}_{\bm \tau}(\text{fold}_{(D,T)}(\bm H) \times_1 \bm S_1^{\dagger} \cdots \times_N \bm S_N^{\dagger}),
\end{align}
where $\bm S_n^{\dagger} := (\bm S_n^T \bm S_n)^{-1} \bm S_n^T $ is a pseudo inverse of $\bm S_n$, $\text{fold}_{(D,T)} := \text{unfold}_{(D,T)}^{-1}$, and $\text{trim}_{\bm \tau} = \text{pad}_{\bm \tau}^{\dagger}$ is a trimming operator for removing  $(\tau_n-1)$ elements at start and end of each mode.
Note that $\mathcal{H}^\dagger \circ \mathcal{H}$ is an identity map, but $\mathcal{H} \circ \mathcal{H}^\dagger$ is not, that is kind of a projection.

\begin{figure}[t]
\centering
\includegraphics[width=0.49\textwidth]{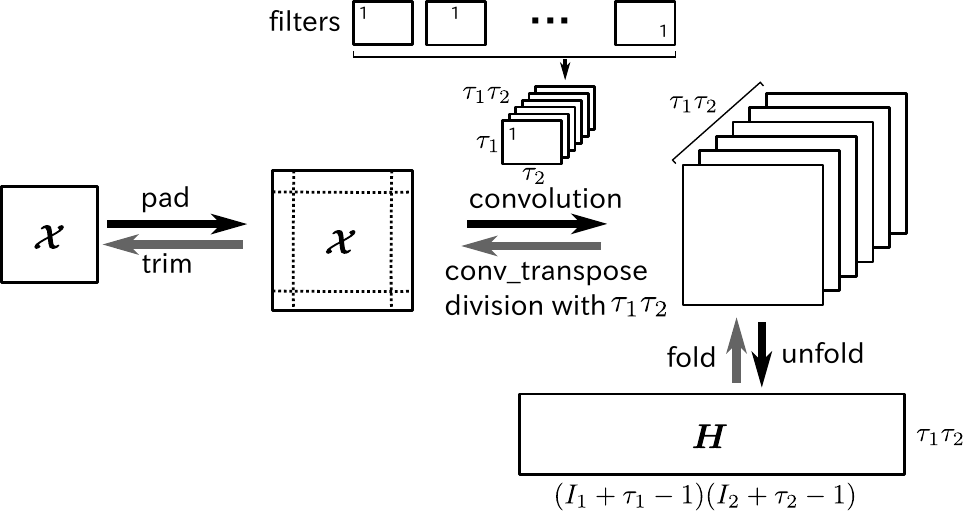}
\caption{Multiway-delay-embedding using convolution ($N=2$).}\label{fig:mdt_conv}
\end{figure}

\subsubsection{Delay embedding using convolution}\label{sec:conv_emb}
Delay embedding and its pseudo inverse can be implemented by using convolution with all one-hot-tensor windows of size $(\tau_1, \tau_2, ..., \tau_N)$.
The one-hot-tensor windows can be given by folding a $D$-dimensional identity matrix $\bm I_D \in \bbR{D \times D}$ into ${\tI}_D \in \bbR{\tau_1 \times \cdots \times \tau_N \times D}$.
Fig.~\ref{fig:mdt_conv} shows a calculation flow of multi-way delay embedding using convolution in a case of $N=2$.
Multi-linear tensor product is replaced with convolution with one-hot-tensor windows.

Pseudo inverse of the convolution with padding is given by its adjoint operation, which is called as the ``transposed convolution'' in some neural network library, with trimming and simple scaling with $D^{-1}$.

\subsection{Definition of low-dimensional manifold}
We consider an AE to define the $r$-dimensional manifold ${\mathcal M}_r$ in ($\prod_n \tau_n$)-dimensional Euclidean space as follows: 
\begin{align}
  &\mathcal{M}_r := \{ \hat{\psi}_r(\bm l) \ | \  \bm l \in \bbR{r} \}, \notag\\ 
  &(\hat{\psi}_r,\hat{\phi}_r) := \mathop{\text{argmin}}_{(\psi_r, \phi_r)} \sum_{t=1}^T || \bm h_t - \psi_r \phi_r(\bm h_t) ||_2^2, \label{eq:def_manifold}
\end{align}
where $\phi_r : \bbR{D} \rightarrow \bbR{r}$ is an encoder, $\psi_r : \bbR{r} \rightarrow \bbR{D}$ is a decoder, and $\hat{\psi}_r \hat{\phi}_r : \bbR{D} \rightarrow \bbR{D}$ is an auto-encoder constructed from $\{\bm h_t\}_{t=1}^T$.
Note that, in general, the use of AE models is a widely accepted approach for manifold learning \cite{hinton2006reducing}.
The properties of the manifold $\mathcal{M}_r$ are determined by the properties of $\phi_r$ and $\psi_r$.
By employing multi-layer perceptrons (neural networks) for $\phi_r$ and $\psi_r$, encoder-decoder may provide a smooth manifold.

\subsection{Problem formulation}
In this section, we combine the conceptual formulation \eqref{eq:concept} and the AE guided manifold constraint to derive a equivalent and more practical optimization problem.
First, we redefine a tensor ${\tX}$ as an output of generator:
\begin{align}
  {\tX} := & {\mathcal H}^{\dagger} [\bm h_1, \bm h_2, ..., \bm h_T], \ \ \text{ where } \bm h_t \in \mathcal{M}_r \notag \\
           = & {\mathcal H}^{\dagger} [\hat{\psi}_r(\bm l_1), \hat{\psi}_r(\bm l_2), ..., \hat{\psi}_r(\bm l_T)], \label{eq:generator}
\end{align}
where $\bm l_t \in \bbR{r}$, and ${\mathcal H}^{\dagger}$ is a pseudo inverse of ${\mathcal H}$.
At this moment, ${\tX}$ is a function of $\{\bm l_t\}_{t=1}^T$, however Hankel structure of matrix $\bm H$ can not be always guaranteed under the unconstrained condition of $\bm l_t$.
For guaranteeing the Hankel structure of matrix $\bm H$, we further transform it as follow:
\begin{align}
  {\tX} := & {\mathcal H}^{\dagger} [\hat{\psi}_r\hat{\phi}_r(\bm g_1), \hat{\psi}_r\hat{\phi}_r(\bm g_2), ..., \hat{\psi}_r\hat{\phi}_r(\bm g_T)], \notag \\
           =& {\mathcal H}^{\dagger} {\mathcal A}_r [\bm g_1, \bm g_2, ..., \bm g_T] \notag \\
           =& {\mathcal H}^{\dagger} {\mathcal A}_r {\mathcal H}({\tZ}), \label{eq:derivation}
\end{align}
where we put ${\mathcal A}_r : \bbR{D \times T} \rightarrow \bbR{D \times T}$ as an operator which auto-encodes each column of a input matrix with $(\hat{\psi}_r,\hat{\phi}_r)$, and $[\bm g_1, \bm g_2, ..., \bm g_T]$ as a matrix, which has Hankel structure and is transformed by Hankelization of some input tensor ${\tZ} \in \bbR{I_1 \times I_2 \times \cdots \times I_N}$.
Note that ${\tZ}$ is the most compact representation for Hankel matrix $[\bm g_1, \bm g_2, ..., \bm g_T]$.
Eq.~\eqref{eq:derivation} describes the MMES network shown in Fig.~\ref{fig:convnet}: ${\mathcal H}$, $\hat{\phi}_r$, $\hat{\psi}_r$ and ${\mathcal H}^{\dagger}$ are respectively corresponding to forward embedding, encoding, decoding, and backward embedding, where encoder and decoder can be defined by multi-layer perceptrons (\ie repetition of linear transformation and non-linear activation).

From this formulation, Problem~\eqref{eq:concept} is transformed as
$ \mathop{\text{minimize}}_{{\tZ}} || {\tY} - {\mathcal F}({\mathcal H}^{\dagger} {\mathcal A}_r {\mathcal H} ({\tZ})) ||_F^2$,
where ${\mathcal A}_r$ is an AE which defines the manifold $\mathcal{M}_r$.
In this study, the AE/manifold is learned from an observed tensor ${\tY}$ itself, thus the optimization problem is finally formulated as
\begin{align}
  \mathop{\text{minimize}}_{{\tZ},{\mathcal A}_r} \ \ & \underbrace{|| {\tY} - {\mathcal F}({\mathcal H}^{\dagger} {\mathcal A}_r {\mathcal H} ({\tZ})) ||_F^2}_{=:\mathcal{L}_{\text{rec}}} \notag \\
                                      &  + \lambda  \underbrace{||{\mathcal H}({\tZ})  - {\mathcal A}_r {\mathcal H} ({\tZ}) ||_F^2}_{=: \mathcal{L}_{\text{AE}}}, \label{eq:cost}
\end{align}
where we refer respectively the first and second terms by a reconstruction loss and an auto-encoding loss, and $\lambda > 0$ is a trade-off parameter for balancing both losses.

\subsection{Design of auto-encoder}
In this section, we discuss how to design the neural network architecture of AE for restricting the manifold ${\mathcal M}_r$.
The simplest way is controlling the value of $r$, and it directly restricts the dimensionality of latent space.
There are many other possibilities: Tikhonov regularization \cite{goodfellow2016deep}, drop-out \cite{gal2016dropout}, denoising auto-encoder \cite{vincent2008extracting,majumdar2018blind}, variational auto-encoder \cite{kingma2013auto}, adversarial auto-encoder \cite{makhzani2015adversarial,creswell2018denoising}, alpha-GAN \cite{rosca2017variational}, and so on.
All methods have some perspective and promise, however the cost is not low.
In this study, we select an attractive and fundamental one: ``denoising auto-encoder''(DAE) \cite{vincent2008extracting}.
The DAE is attractive because it has a strong relationship with Tikhonov regularization \cite{bishop1995training}, and decreases the entropy of data \cite{sonoda2017transportation}.
Furthermore, learning with noise is also employed in the deep image prior.

\begin{figure}[t]
\centering
\includegraphics[width=0.35\textwidth]{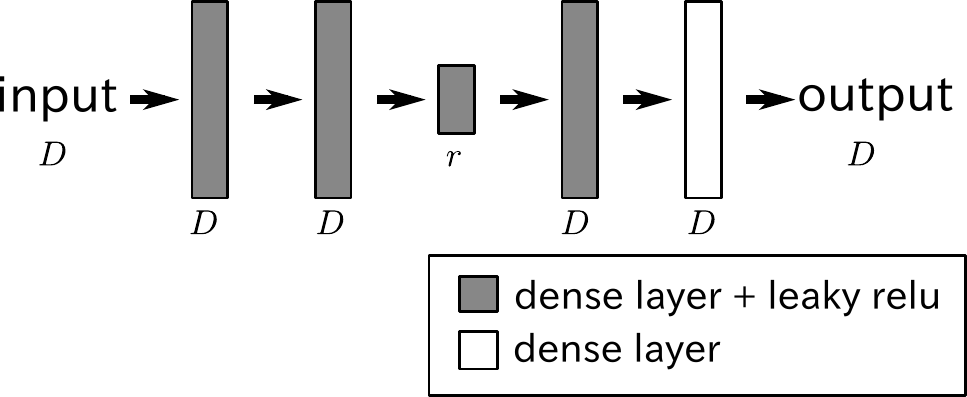}
\caption{An example of architecture of auto-encoder.}\label{fig:dae}
\end{figure}

Finally, we designed an auto-encoder with controlling the dimension $r$ and the standard deviation $\sigma$ of additive zero-mean Gaussian noise.
Fig.~\ref{fig:dae} shows the illustration of examples of architecture of auto-encoder which we used in this study.
The sizes of hidden variables affect the representation ability for image reconstruction.

\begin{figure}[t]
\centering
\includegraphics[width=0.49\textwidth]{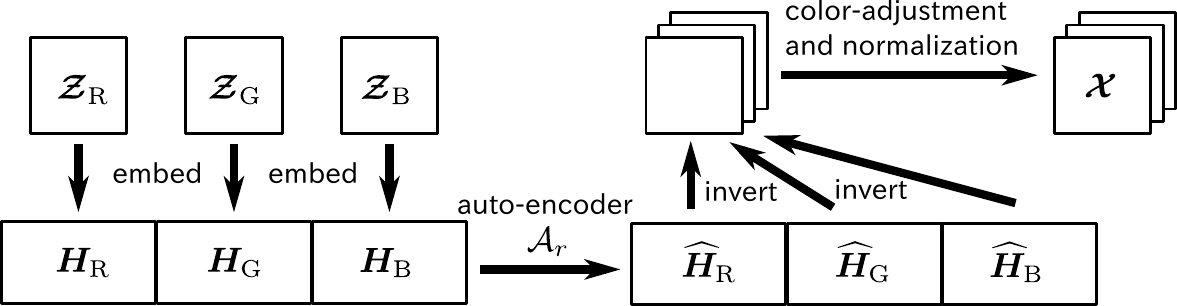}
\caption{Generator network in a case of color-image recovery.}\label{fig:color_image_case}
\end{figure}

\subsection{Optimization}
Optimization problem~\eqref{eq:cost} consists of two terms: a reconstruction loss, and an auto-encoding loss.
Hyperparameter $\lambda$ is set to balance both losses.
Basically, $\lambda$ should be large because auto-encoding loss should be zero.
However, very large $\lambda$ prohibits minimizing the reconstruction loss, and may lead to local optima.
Therefore, we adjust gradually the value of $\lambda$ in the optimization process.

Algorithm~\ref{alg:opt} shows an optimization algorithm for tensor reconstruction and/or enhancement.
For AE learning, we employs a strategy of denoising-auto-encoder. 
Adaptation of $\lambda$ is just an example, and it can be modified appropriately with data/tasks.
Here, the trade-off parameter $\lambda$ is adjusted for keeping $\mathcal{L}_{\text{rec}} > \mathcal{L}_{\text{AE}}$, but for no large gap between both losses.
By exploiting the convolutional structure of $\mathcal{H}$ and $\mathcal{H}^\dagger$, the calculation flow of $\mathcal{L}_{\text{rec}}$ and $\mathcal{L}_{\text{AE}}$ can be easily implemented by using neural network libraries such as \texttt{TensorFlow}.
We employed \texttt{Adam} \cite{kingma2014adam} optimizer for updating $({\tZ}, \mathcal{A}_r)$.

\begin{algorithm}[t]
\caption{Optimization algorithm for tensor reconstruction}\label{alg:opt}
\begin{algorithmic}[0]
  \STATE {\bf input}: ${\tY} \in \bbR{J_1 \times \cdots \times J_N}$ (corrupted tensor), $\mathcal{F}$, $\bm \tau$, $r$, $\sigma$;
  \STATE {\bf initialize}: ${\tZ} \in \bbR{I_1 \times \cdots \times I_N}$, auto-encoder $\mathcal{A}_r$, $\lambda = 5.0$;
  \REPEAT
    \STATE $\bm H \leftarrow \mathcal{H}({\tZ}) \in \bbR{D \times T}$ with $\bm\tau$;
    \STATE generate noise $\bm E \in \bbR{D \times T}$ with $\sigma$;
    \STATE $\mathcal{L}_{\text{AE}} \leftarrow ||\bm H - \mathcal{A}_r(\bm H + \bm E)||_F^2$;
    \STATE $\mathcal{L}_{\text{rec}} \leftarrow \frac{1}{D}|| {\tY} - {\mathcal F}({\mathcal H}^{\dagger} \mathcal{A}_r(\bm H + \bm E)) ||_F^2 $;
    \STATE update $({\tZ}, \mathcal{A}_r)$ by \texttt{Adam} for $\mathcal{L}_{\text{rec}}+\lambda\mathcal{L}_{\text{AE}}$;
    \STATE {\bf if} $\mathcal{L}_{\text{rec}} < \mathcal{L}_{\text{AE}}$ {\bf then} $\lambda \leftarrow 1.1 \lambda$; {\bf else} $\lambda \leftarrow 0.99 \lambda$;
  \UNTIL{converge}
  \STATE {\bf output}: $\widehat{{\tX}} = {\mathcal H}^{\dagger} \mathcal{A}_r {\mathcal H}({\tZ}) \in \bbR{I_1 \times \cdots \times I_N}$ (reconstructed tensor);
\end{algorithmic}
\end{algorithm}

\subsection{A special setting for color-image recovery}
In case of multi-channel or color image recovery case, we use a special setting of generator network because spacial pattern of individual channels are similar and the patch-manifold can be shared.
Fig.~\ref{fig:color_image_case} shows an illustration of the auto-encoder shared version of MMES in a case of color image recovery.
In this case, we put three channels of input and each channel input is embedded, independently.
Then, three block Hankel matrices are concatenated, and auto-encoded simultaneously.
Inverted three images are stacked as a color-image (third-order tensor), and finally color-transformed.
The last color-transform can be implemented by convolution layer with kernel size (1,1), and it is also optimized as parameters.
It should be noted that the input three channels are not necessary to correspond to RGB, but it would be optimized as some compact color-representation.

\section{Experiments}
Here, we show the extensive experimental results to demonstrate the close similarity and some slight differences between DIP and MMES.
First, toy examples with a time-series signal and a gray-scale image were recovered by the proposed method to show its basic behaviors.
Second, hyper-parameter sensitivity was demonstrated to get a sense for adjusting parameters, and to show the effects of denoising auto-encoder.
Third, the phenomenon of noise impedance in MMES was demonstrated compared with DIP.
Finally, we show the results by comparison with DIP and other selective methods on color-image inpainting, super-resolution, deconvolution, and denoising tasks.


\begin{figure}[t]
\centering
\includegraphics[width=0.49\textwidth]{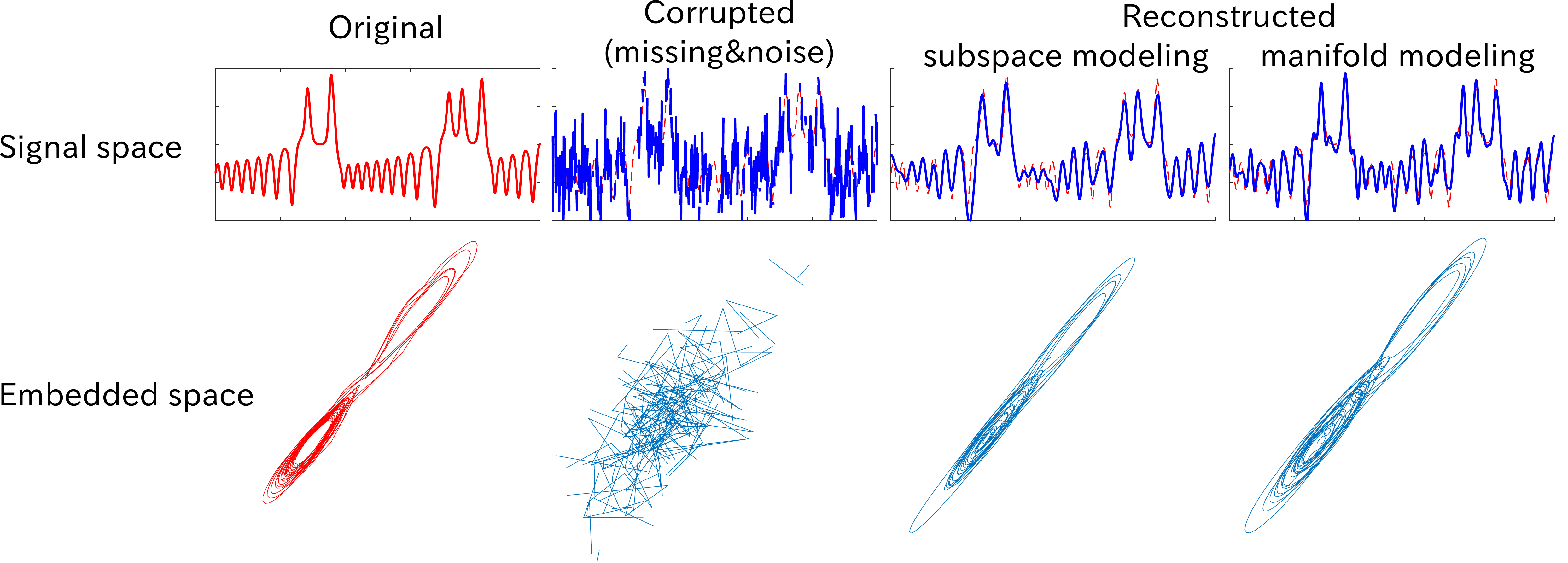}
\caption{Time series signal recovery of subspace and manifold models in embedded space.}\label{fig:signal}
\end{figure}
\begin{figure}[t]
\centering
\includegraphics[width=0.49\textwidth]{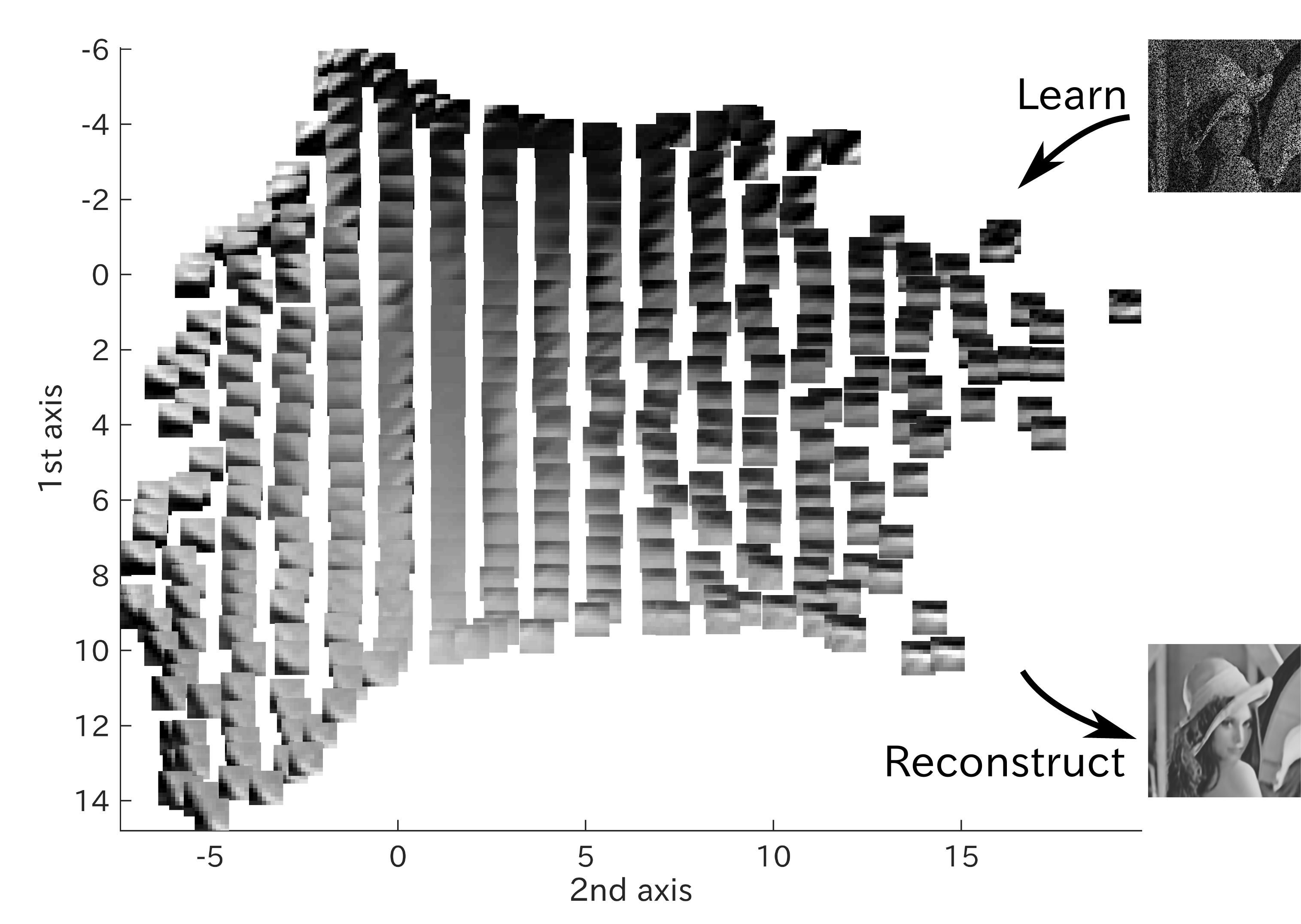}
\caption{Two-dimensional (8,8)-patch manifold learned from a 50\% missing gray-scale image of `Lena'.}\label{fig:patch_manifold}
\end{figure}

\subsection{Toy examples}
In this section, we apply the proposed method into a toy example of signal recovery.
Fig.~\ref{fig:signal} shows a result of this experiment.
A one-dimensional time-series signal is generated from Lorentz system, and corrupted by additive Gaussian noise, random missing, and three block occlusions.
The corrupted signal was recovered by the subspace modeling \cite{yokota2018missing}, and the proposed manifold modeling in embedded space.
Window size of delay-embedding was $\tau=64$, the lowest dimension of auto-encoder was $r=3$, and additive noise standard deviation was set to $\sigma = 0.05$.
Manifold modeling catched the structure of Lorentz attractor much better than subspace modeling.

Fig.~\ref{fig:patch_manifold} visualizes a two-dimensional $(8,8)$-patch manifold learned by the proposed method from a 50\% missing gray-scale image of `Lena'.  For this figure, we set $\bm\tau=[8,8]$, $r=2$, $\sigma=0.05$.
Similar patches are located near each other, and the smooth change of patterns can be observed.
It implies the relationship between non-local similarity based methods \cite{buades2005non,dabov2007image,gu2014weighted,zhang2014group}, and the manifold modeling (\ie DAE) plays a key role of ``patch-grouping'' in the proposed method.
The difference from the non-local similarity based approach is that the manifold modeling is ``global'' rather than ``non-local'' which finds similar patches of the target patch from its neighborhood area.

\subsubsection{Optimization behavior}
\begin{figure}[t]
\centering
\includegraphics[width=0.45\textwidth]{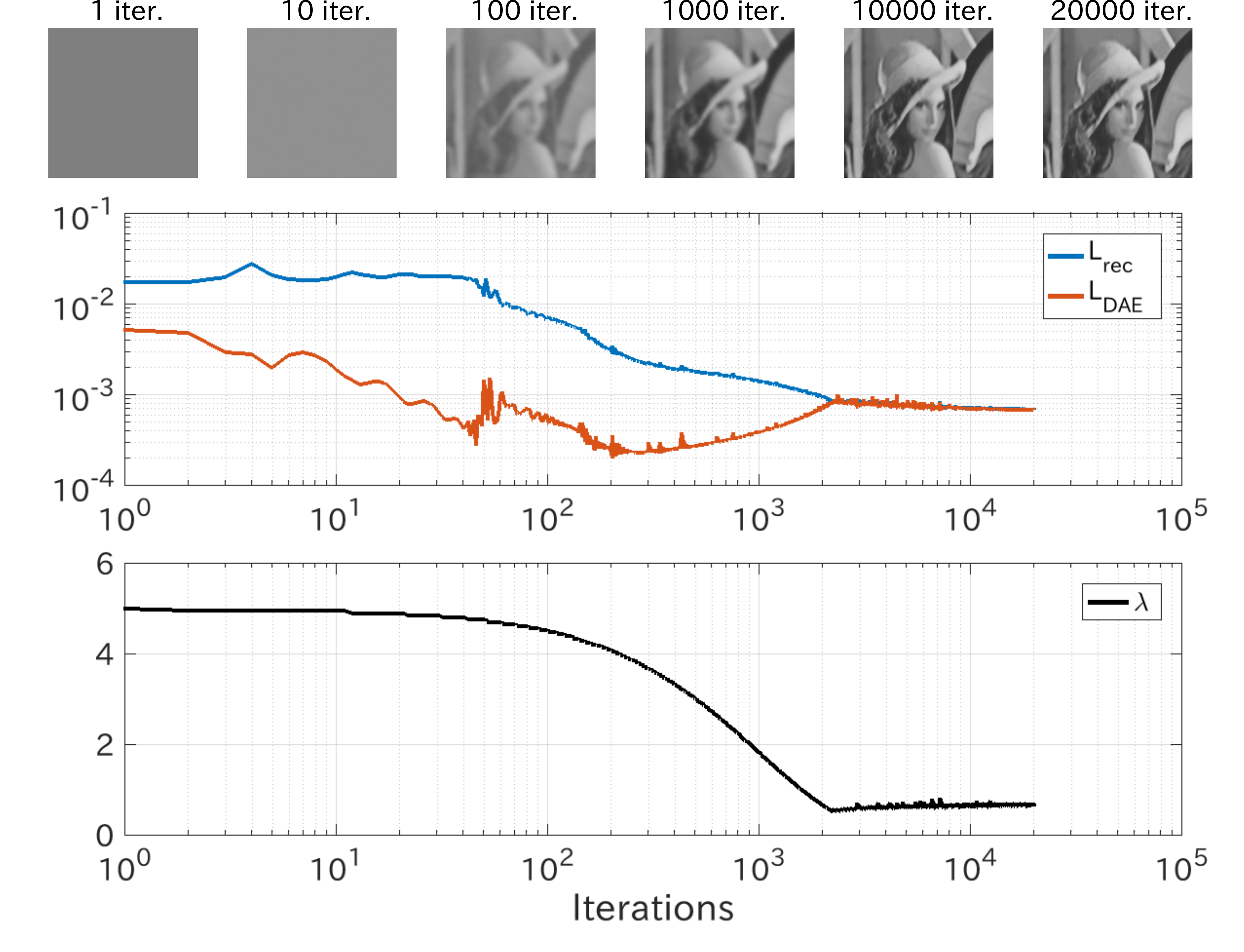}
\caption{Optimization behavior.}\label{fig:opt}
\end{figure}

For this experiment, we recovered 50\% missing gray-scale image of `Lena'.
We stopped the optimization algorithm after 20,000 iterations.
Learning rate was set as 0.01, and we decayed the learning rate with 0.98 every 100 iterations.
$\lambda$ was adapted by Algorithm~\ref{alg:opt} every 10 iterations.
Fig.~\ref{fig:opt} shows optimization behaviors of reconstructed image, reconstruction loss $\mathcal{L}_{\text{rec}}$, auto-encoding loss $\mathcal{L}_{\text{DAE}}$, and trade-off coefficient $\lambda$.
By using trade-off adjustment, the reconstruction loss and the auto-encoding loss were intersected around 1,500 iterations, and both losses were jointly decreased after the intersection point.

\subsection{Hyper-parameter sensitivity}

\begin{figure}[t]
\centering
\includegraphics[width=0.45\textwidth]{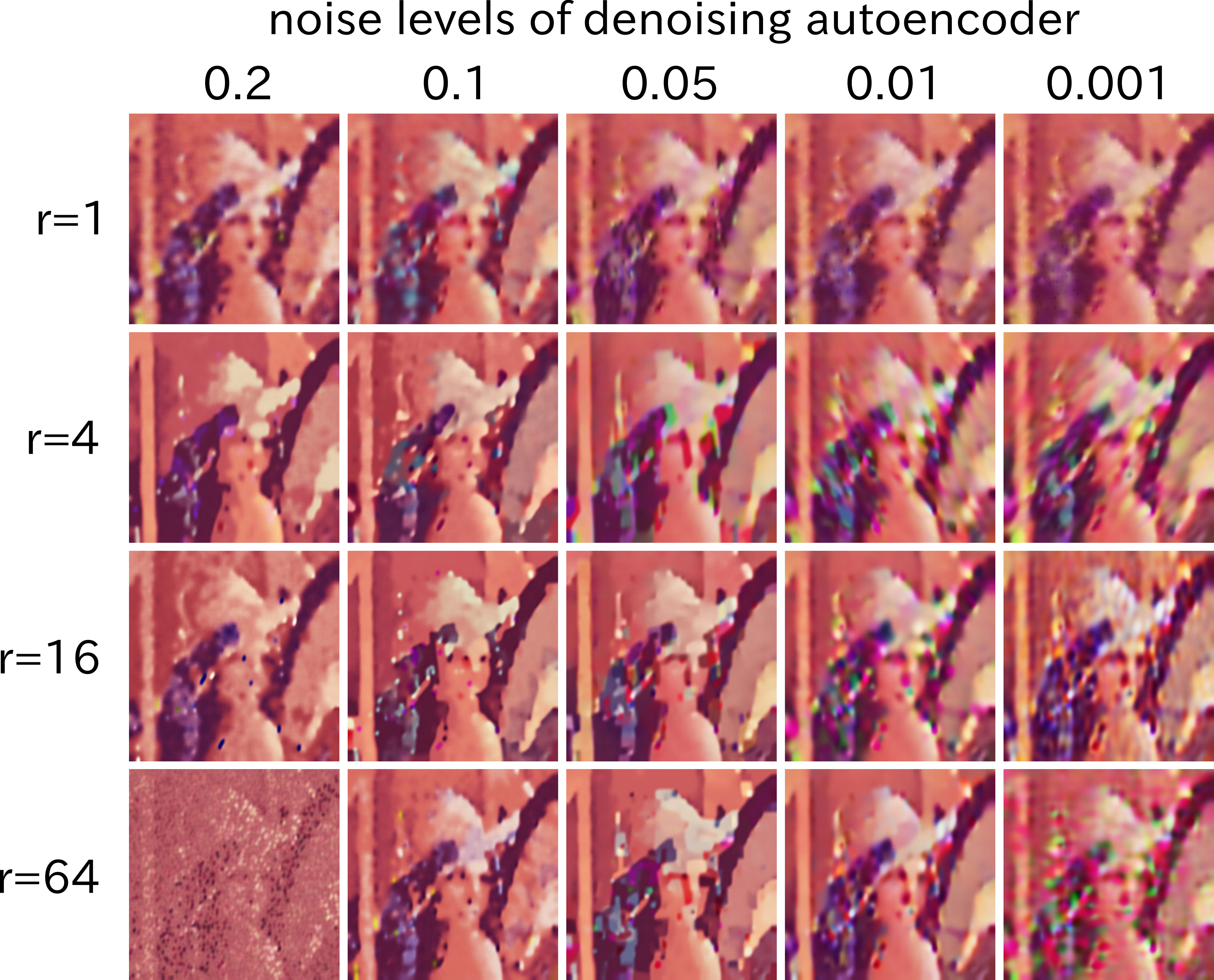}
\caption{Performance of reconstruction of color image of `Lena' with 99\% pixels missing for various parameter setting.}\label{fig:r_sigma}
\vspace*{5mm}
\includegraphics[width=0.4\textwidth]{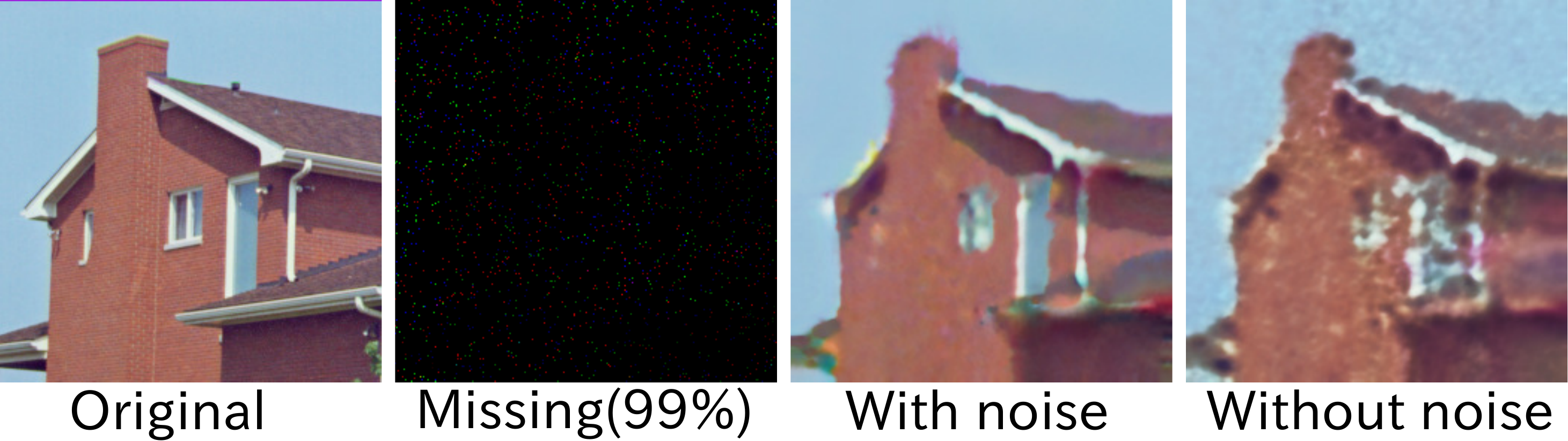}
\caption{Reconstruction of `home' image by training with/without noise in deep image prior.}\label{fig:DIP_noise}
\vspace*{5mm}
\centering
\includegraphics[width=0.49\textwidth]{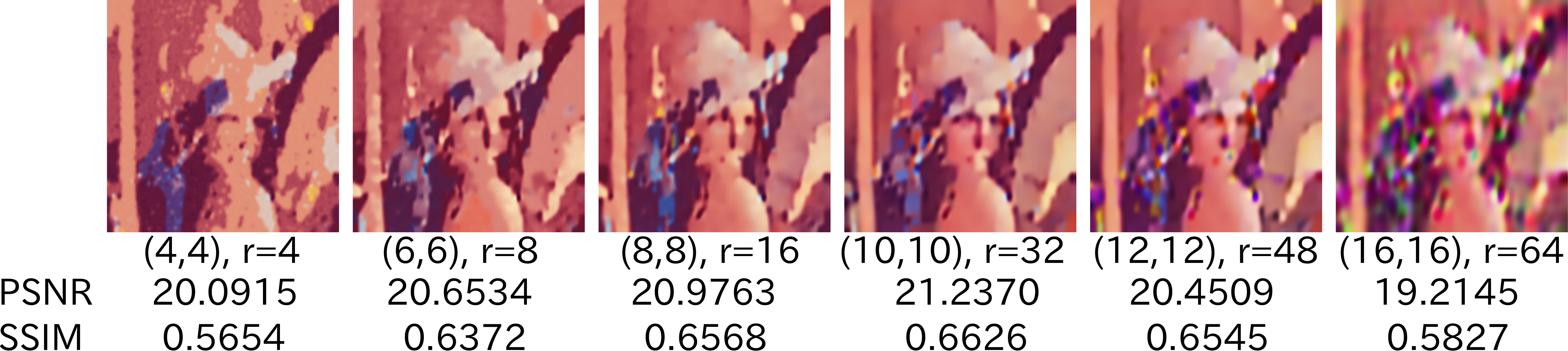}
\caption{Reconstruction of `Lena' image for various patch sizes $\bm\tau$.}\label{fig:var_tau}
\end{figure}

We evaluate the sensitivity of MMES with three hyper-parameters: $r$, $\sigma$, and $\bm\tau$.
First, we fixed the patch-size as $(8,8)$, and dimension $r$ and noise standard deviation $\sigma$ were varied.
Fig.~\ref{fig:r_sigma} shows the reconstruction results of a 99\% missing image of `Lena' by the proposed method with different settings of $(r,\sigma)$.
The proposed method with very low dimension ($r=1$) provided blurred results, and the proposed method with very high dimension ($r=64$) provided results which have many peaks.
Furthermore, some appropriate noise level ($\sigma=0.05$) provides sharp and clean results.
For reference, Fig.~\ref{fig:DIP_noise} shows the difference of DIP optimized with and without noise.
From both results, the effects of learning with noise can be confirmed.

Next, we fixed the noise level as $\sigma=0.05$, and the patch-size were varied with some values of $r$.
Fig.~\ref{fig:var_tau} shows the results with various patch-size settings for recovering a 99\% missing image.
The patch sizes $\bm\tau$ of (8,8) or (10,10) were appropriate for this case.
Patch size is very important because it depends on the variety of patch patterns.  If patch size is too large, then patch variations might expand and the structure of patch-manifold is complicated.  By contrast, if patch size is too small, then the information obtained from the embedded matrix $\bm H$ is limited and the reconstruction becomes difficult in highly missing cases.  The same problem might be occurred in all patch-based image reconstruction methods \cite{buades2005non,dabov2007image,gu2014weighted,zhang2014group}.
However, good patch sizes would be different for different images and types/levels of corruption, and the estimation of good patch size is an open problem.
Multi-scale approach \cite{yair2018multi} may reduce a part of this issue but the patch-size is still fixed or tuned as a hyper-parameter.

\subsection{Noise impedance of MMES}\label{sec:noise_impedance}
\begin{figure*}[t]
  \centering
  \includegraphics[width=0.85\textwidth]{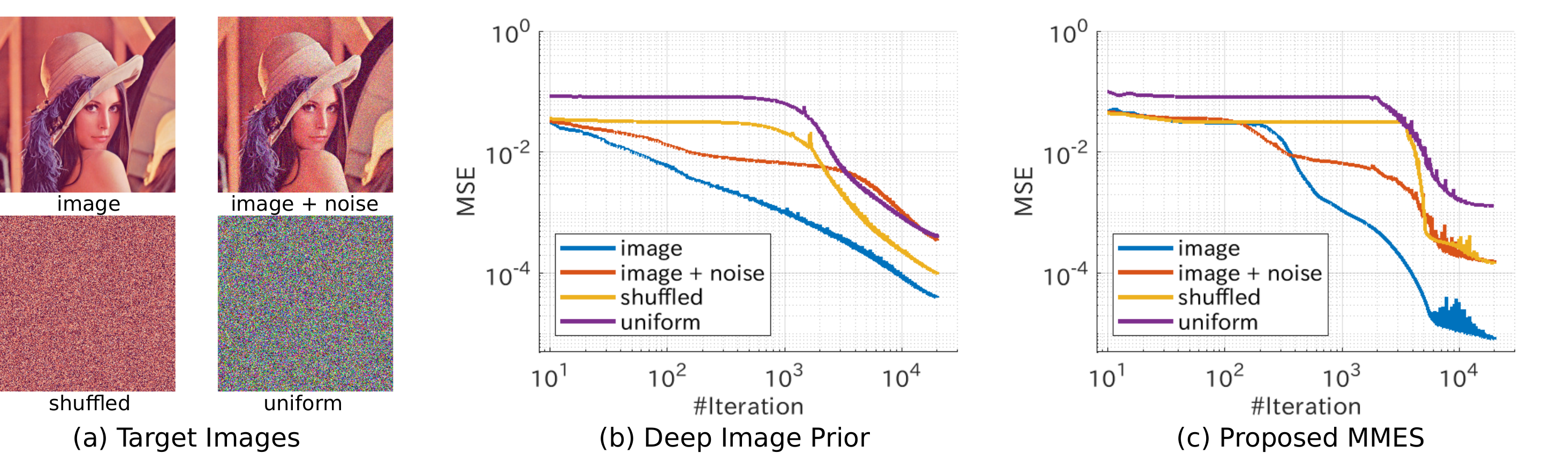}
  \caption{Optimization behaviors of DIP and MMES for different target images. (a) Four target images. (b) and (c) show optimization behaviors of DIP and MMES. Natural image is optimized faster than noisy images in both DIP and MMES.}\label{fig:impedance}
\end{figure*}

Here, we reproduce the demonstration of noise impedance in \cite{ulyanov2018deep} with the proposed MMES.
Four target color-images: a natural image, a natural image with noise $ \sim \mathcal{N}(0,20^2)$, a pixel-shuffled image, and a uniform noise were prepared as shown in Fig.~\ref{fig:impedance}(a) and were reconstructed by DIP and MMES.
In both methods, we recorded the mean squares error between the target and reconstructed images in each iteration.
Learning rate was set as 0.01, and we decayed the learning rate with 0.98 every 100 iterations.
In MMES, the trade-off parameter $\lambda$ was dynamically adjusted so that the auto-encoding loss $\mathcal{L}_{\text{DAE}}$ does not exceed some small value.

Fig.~\ref{fig:impedance}(b) and (c) show the optimization behaviors of DIP and MMES for reconstructing four target images.
First, a natural image was reconstructed earliest in both DIP and MMES.
Optimization of the noisy image in both DIP and MMES are slower than the noise-free one, and both curves were similar.
The curves of shuffled and uniform images of MMES were also similar to DIP.
It implies that the MMES has noise impedance.


\subsection{Comparisons}
Here, we show experimental results of performance comparison in four tasks: tensor completion, super-resolution, deconvolution, and denoising.

\begin{figure}[t]
\centering
\includegraphics[width=0.49\textwidth]{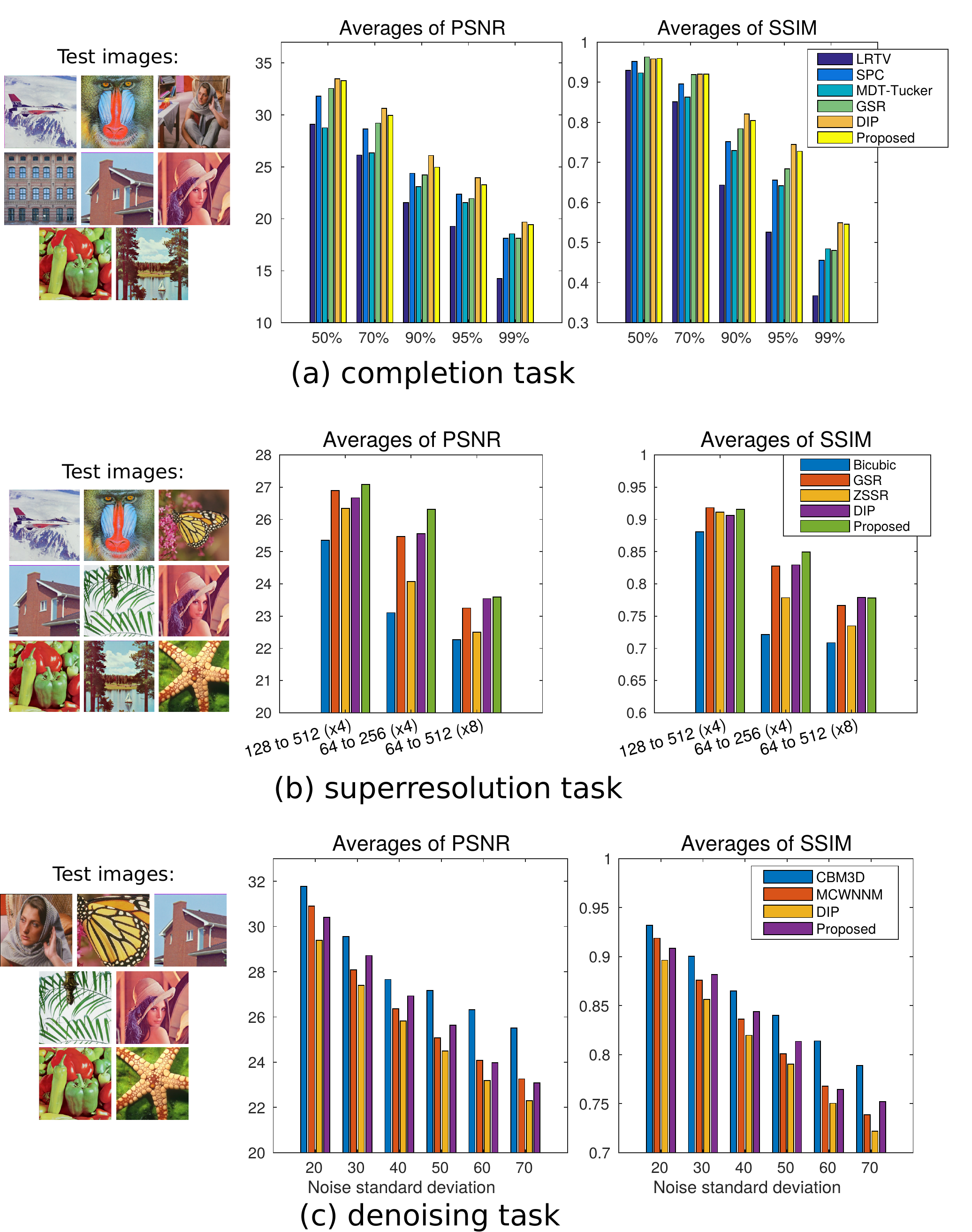}
\caption{Comparison of averages of PSNR and SSIM for eight color image completion with various missing rates (from 50\% to 99\% missing pixels).}\label{fig:psnr_bar}
\end{figure}

\begin{figure*}[t]
\centering
\includegraphics[width=0.95\textwidth]{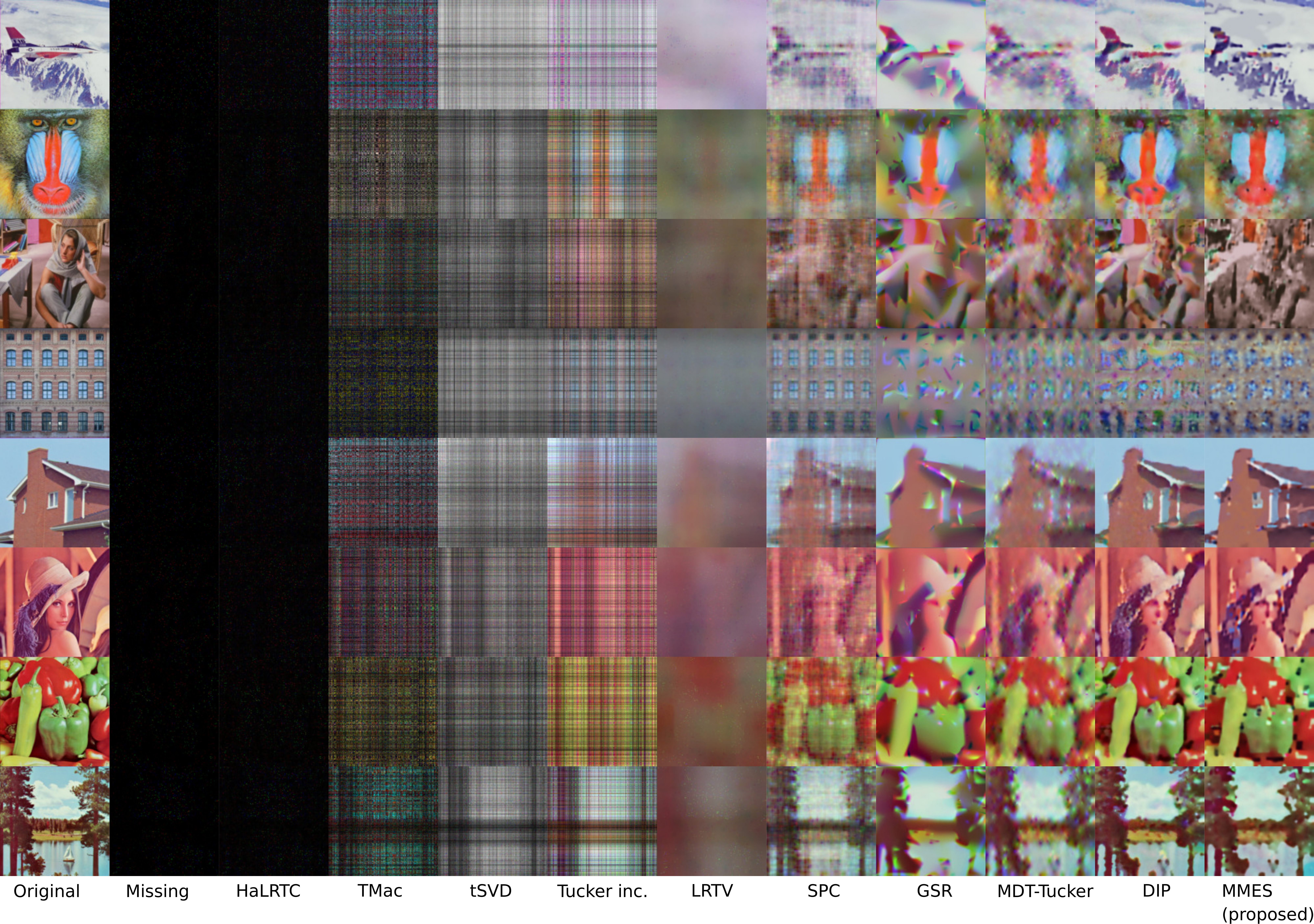}
\caption{Completion results from images with 99\% missing pixels by HaLTRC \cite{liu2013tensor}, TMac \cite{xu2015parallel}, tSVD \cite{zhang2014novel}, Tucker inc. \cite{yokota2018missing}, LRTV \cite{yokota2017simultaneous}, SPC \cite{yokota2016smooth}, GSR \cite{zhang2014group}, MDT-Tucker \cite{yokota2018missing}, DIP \cite{ulyanov2018deep} and the proposed MMES.}\label{fig:compare}
\vspace{5mm}
\centering
\includegraphics[width=0.85\textwidth]{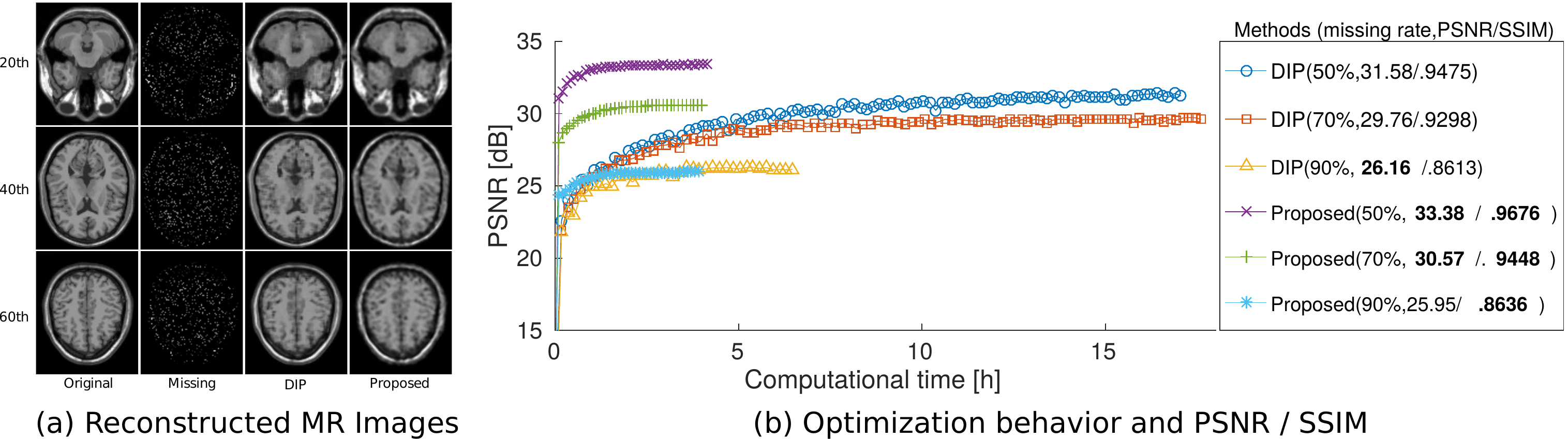}
\caption{Results of MRI completion: (a) Illustration of MRI reconstructed from 90\% missing tensor, and (b) optimization behaviors of PSNR with final values of PSNR/SSIM by DIP and proposed MMES.}\label{fig:MRI}
\end{figure*}

\subsubsection{Color image completion, especially for extremely high number of missing pixels}\label{sec:completion}
In this section, we compared performance of the proposed method with several selected unsupervised tensor completion/image inpainting methods: low-rank tensor completion (HaLRTC) \cite{liu2013tensor}, parallel low-rank matrix factorization (TMac) \cite{xu2015parallel}, tubal nuclear norm regularization (tSVD) \cite{zhang2014novel},  Tucker decomposition with rank increment (Tucker inc.) \cite{yokota2018missing}, low-rank and total-variation (LRTV) regularization\footnote{For LRTV, the MATLAB software was downloaded from \url{https://sites.google.com/site/yokotatsuya/home/software/lrtv_pds}} \cite{yokota2017simultaneous,yokota2019simultaneous}, smooth PARAFAC tensor completion (SPC)\footnote{For SPC, the MATLAB software was downloaded from \url{https://sites.google.com/site/yokotatsuya/home/software/smooth-parafac-decomposition-for-tensor-completion}.} \cite{yokota2016smooth}, GSR\footnote{For GSR, each color channel was recovered, independently, using the MATLAB software downloaded from \url{https://github.com/jianzhangcs/GSR}.} \cite{zhang2014group}, multi-way delay embedding based Tucker modeling (MDT-Tucker)\footnote{For MDT-Tucker, the MATLAB software was downloaded from \url{https://sites.google.com/site/yokotatsuya/home/software/mdt-tucker-decomposition-for-tensor-completion}.} \cite{yokota2018missing}, and DIP\footnote{For DIP, we implemented by ourselves in Python with \texttt{TensorFlow}.} \cite{ulyanov2018deep}.

For this experiments, hyper-parameters of all methods were tuned manually to perform the best peak-signal-to-noise ratio (PSNR) and for structural similarity (SSIM), although it would not be perfect.
For DIP, we did not try the all network structures with various kernel sizes, filter sizes, and depth.  We just employed ``default architecture'', which the details are available in supplemental material\footnote{\url{https://dmitryulyanov.github.io/deep_image_prior}} of \cite{ulyanov2018deep}, and employed the best results at the appropriate intermediate iterations in optimizations based on the value of PSNR. 
For the proposed MMES method, we adaptively selected the patch-size $\bm\tau$, and dimension $r$.
Table~\ref{tab:para} shows parameter settings of $\bm\tau=[\tau, \tau]$ and $r$ for MMES.  Noise level of denoising auto-encoder was set as $\sigma=0.05$ for all images.
For auto-encoder, same architecture shown in Fig.~\ref{fig:dae} was employed.
Initial learning rate of Adam optimizer was 0.01 and we decayed the learning rate with 0.98 every 100 iterations.
The optimization was stopped after 20,000 iterations for each image.

Fig.~\ref{fig:psnr_bar}(a) shows the eight test images and averages of PSNR and SSIM for various missing ratio \{50\%, 70\%, 90\%, 95\%, 99\%\} and for selective competitive methods.
The proposed method is quite competitive with DIP.
Fig.~\ref{fig:compare} shows the illustration of results.
The 99\% of randomly selected voxels are removed from 3D (256,256,3)-tensors, and the tensors were recovered by various methods.
Basically low-rank priors (HaLRTC, TMac, tSVD, Tucker) could not recover such highly incomplete image.
In piecewise smoothness prior (LRTV), over-smoothed images were reconstructed since the essential image properties could not be captured.
There was a somewhat jump from them by SPC (i.e., smooth prior of basis functions in low-rank tensor decomposition).
MDT-Tucker further improves it by exploiting the shift-invariant multi-linear basis.
GSR nicely recovered the global pattern of images but details were insufficient. 
Finally, the reconstructed images by DIP and MMES recovered both global and local patterns of images.


\begin{table}[t]
\caption{Parameter settings for MMES in image completion experiments}\label{tab:para}
\centering
\renewcommand{\tabcolsep}{0.05cm}
{
\begin{tabular}{l | l l l l l l l l}\hline
 $(\tau, r)$ & airplane & baboon & barbara & facade & house  & lena    & peppers  & saiboat \\ \hline
50 \%        & (16,4)   & (10,4) & (6,4)   & (10,4) & (16,4) & (6,4)   & (6,4)    & (6,4)   \\
70 \%        & (16,4)   & (10,4) & (6,4)   & (16,4) & (16,4) & (6,4)   & (16,4)   & (6,4)   \\
90 \%        & (16,4)   & (4,8)  & (6,4)   & (16,4) & (16,4) & (8,4)   & (16,4)   & (4,4)   \\
95 \%        & (16,4)   & (4,6)  & (6,4)   & (16,4) & (16,4) & (6,8)   & (16,4)   & (6,8)   \\
99 \%        & (8,32)   & (4,4)  & (6,4)   & (4,1)  & (8,16) & (10,32) & (8,8)    & (6,4)   \\ \hline
\end{tabular}
}
\end{table}



\subsubsection{Volumetric/3D image/tensor completion}
In this section, we show the results of MR-image/3D-tensor completion problem.
The size of MR image is (109,91,91).
We randomly remove 50\%, 70\%, and 90\% voxels of the original MR-image and recover the missing MR-images by the proposed method and DIP.
For DIP, we implemented the 3D version of {\em default architecture} in \texttt{TensorFlow}, but the number of filters of shallow layers were slightly reduced because of the GPU memory constraint.
For the proposed method, 3D patch-size was set as $\tau=[4,4,4]$, the lowest dimension was $r=6$, and noise level was $\sigma=0.05$.  Same architecture shown in Fig.~\ref{fig:dae} was employed.

Fig.~\ref{fig:MRI} shows reconstruction results and behavior of PSNR with final value of PSNR/SSIM in this experiment.
From the values of PSNR and SSIM, the proposed MMES outperformed DIP in low-rate missing cases, and it is quite competitive in highly missing cases.
The some degradation of DIP might be occurred by the insufficiency of filter sizes since much more filter sizes would be required for 3D ConvNet than 2D ConvNet.
Moreover, computational times required for our MMES were significantly shorter than that of DIP.
%
\begin{figure*}[t]
  \centering
  \includegraphics[width=0.99\textwidth]{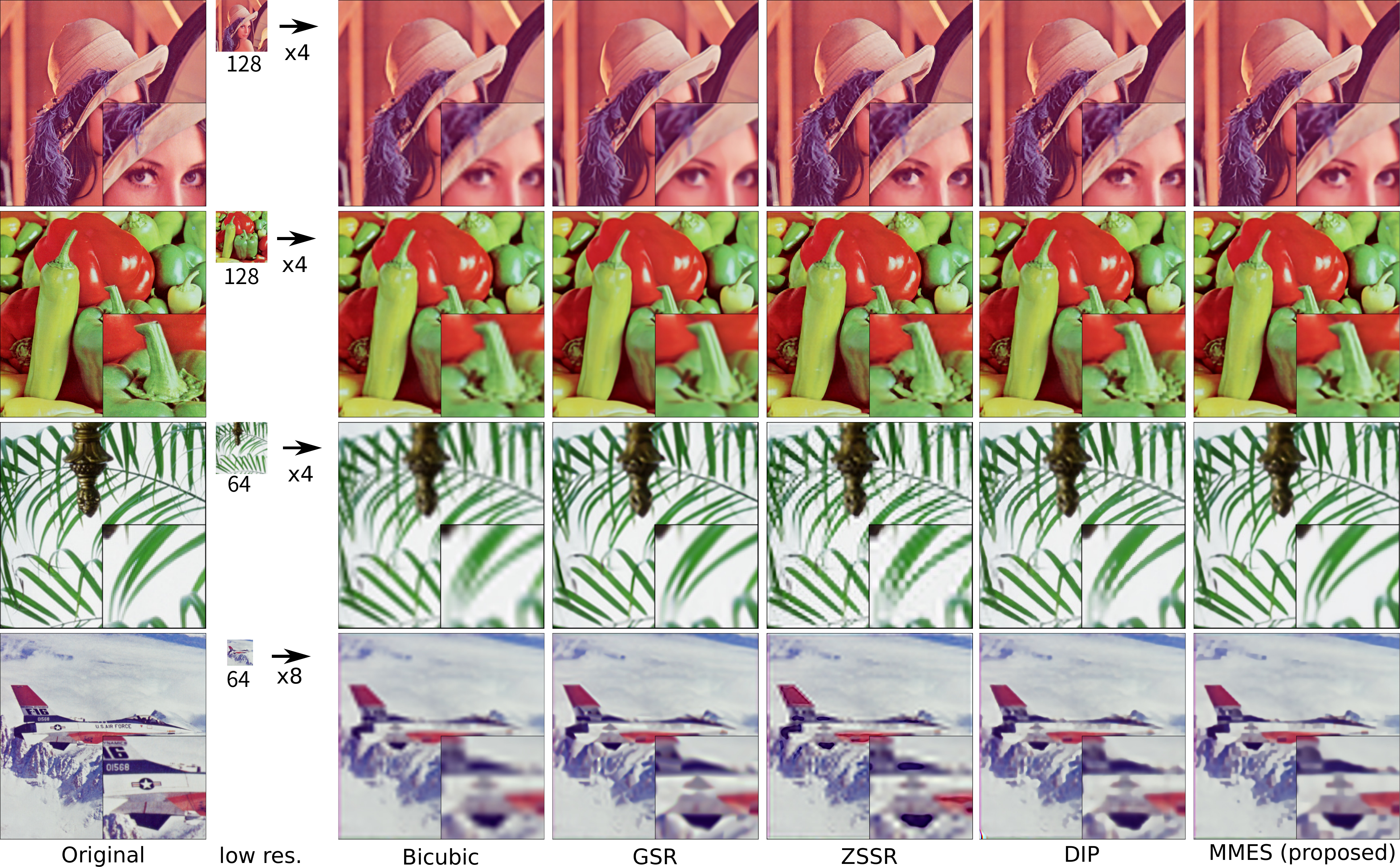}
  \caption{Super-resolution results: The first and second lines `Starfish' and `Leaves' were up-scaled from (64,64,3) to (256,256,3), the third line `Monarch' was up-scaled from (128,128,3) to (512,512,3), and the fourth line `Airplane' was up-scaled from (64,64,3) to (512,512,3).}\label{fig:SR}
\end{figure*}

\subsubsection{Color image superresolution}
In this section, we compare performance of the proposed method with several selected unsupervised image super-resolution methods: bicubic interpolation, GSR\footnote{For GSR, each color channel was recovered, independently, using the MATLAB software downloaded from \url{https://github.com/jianzhangcs/GSR}.  We slightly modified its MATLAB code for applying it to super-resolution task.} \cite{zhang2014group}, ZSSR\footnote{For ZSSR, software was downloaded from \url{https://github.com/assafshocher/ZSSR}.  We set the same Lanczos2 kernel for this super-resolution task.} and DIP \cite{ulyanov2018deep}.

In this experiments, DIP was conducted with the best number of iterations from \{1000, 2000, 3000, ..., 9000\}.
For four times (x4) up-scaling in MMES, we set $\tau=6$, $r=32$, and $\sigma=0.1$.
For eight times (x8) up-scaling in MMES, we set $\tau=6$, $r=16$, and $\sigma=0.1$.
For all images in MMES, the architecture of auto-encoder consists of three hidden layers with sizes of $[8\tau^2, r, 8\tau^2]$.
We assumed the same Lanczos2 kernel for down-sampling system for all super-resolution methods.

Fig.~\ref{fig:psnr_bar}(b) shows the nine test images and averages of PSNR and SSIM for three super-resolution settings.
We used three (256,256,3) color images, and six (512,512,3) color images.
Super resolution methods scaling up them from four or eight times down-scaled images of them.
According to this quantitative evaluation, bicubic interpolation was clearly worse than others.
Basically, GSR, DIP, and MMES were very competitive.
In detail, DIP was slightly better than GSR, and the proposed MMES was slightly better than DIP.

Fig.~\ref{fig:SR} shows selected high resolution images reconstructed by four super-resolution methods.
In general, bicubic method reconstructed blurred images and these were visually worse than others.
GSR results had smooth outlines in all images, but these were slightly blurred.
ZSSR was weak for very low-resolution images.
DIP reconstructed visually sharp images but these images had jagged artifacts along the diagonal lines.
The proposed MMES reconstructed sharp and smooth outlines.

\begin{figure*}
\centering
\includegraphics[width=0.99\textwidth]{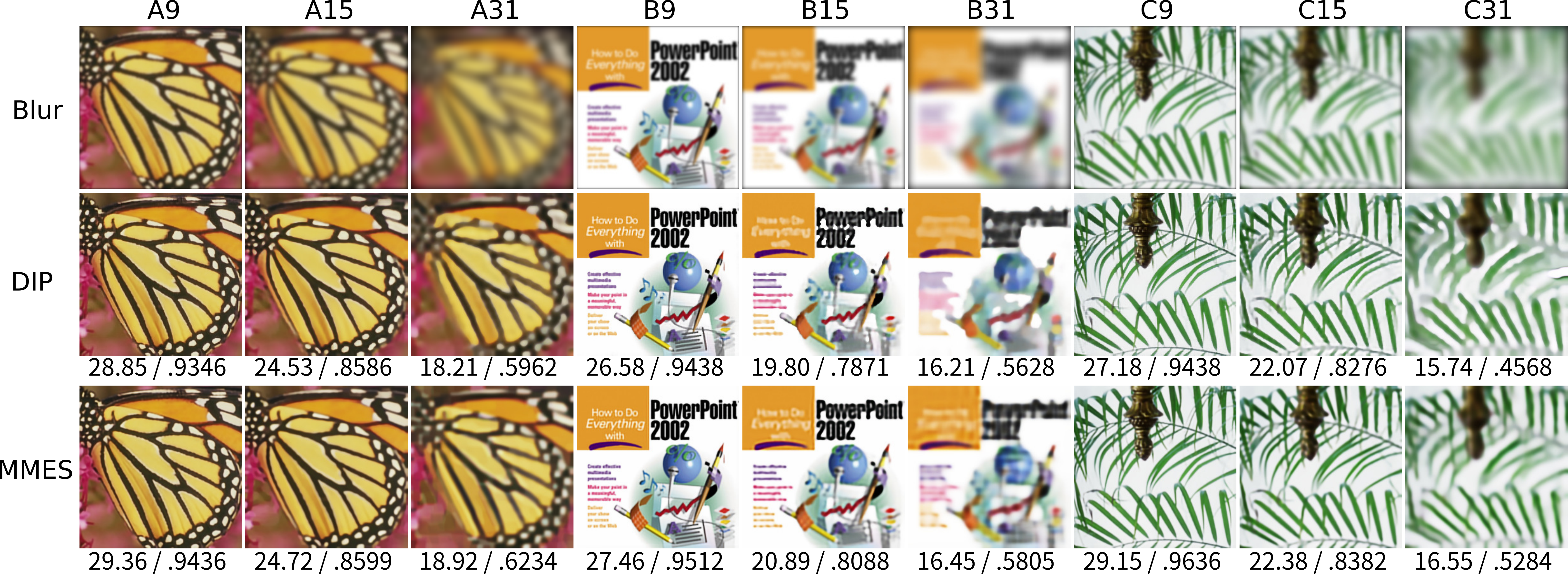}
\caption{Comparison of our approach with DIP for deconvolution/deblurring task. Three color images were blurred by three Gaussian windows of different sizes. These were recovered by the DIP and the proposed MMES.}\label{fig:deconv}
\end{figure*}

\subsubsection{Color image deconvolution}\label{sec:deconv}
In this section, we compare the proposed method with DIP for image deconvolution/deblurring task.  Three (256,256,3) color images are prepared and blurred by using three different Gaussian filters.
For DIP we choose the best early stopping timing from \{1000, 2000, ..., 10000\} iterations.
For MMES, we employed the fixed AE structure as $[32\tau^2, r, 32\tau^2]$, and parameters as $\tau=4$, $r=16$, and $\sigma=0.01$ for all nine cases.
Fig.~\ref{fig:deconv} shows the reconstructed deblurring images by DIP and MMES with these PSNR and SSIM values.
We can see that the similarity of the methods qualitatively and quantitatively.

%
%

\begin{figure*}[t]
  \centering
  \includegraphics[width=0.99\textwidth]{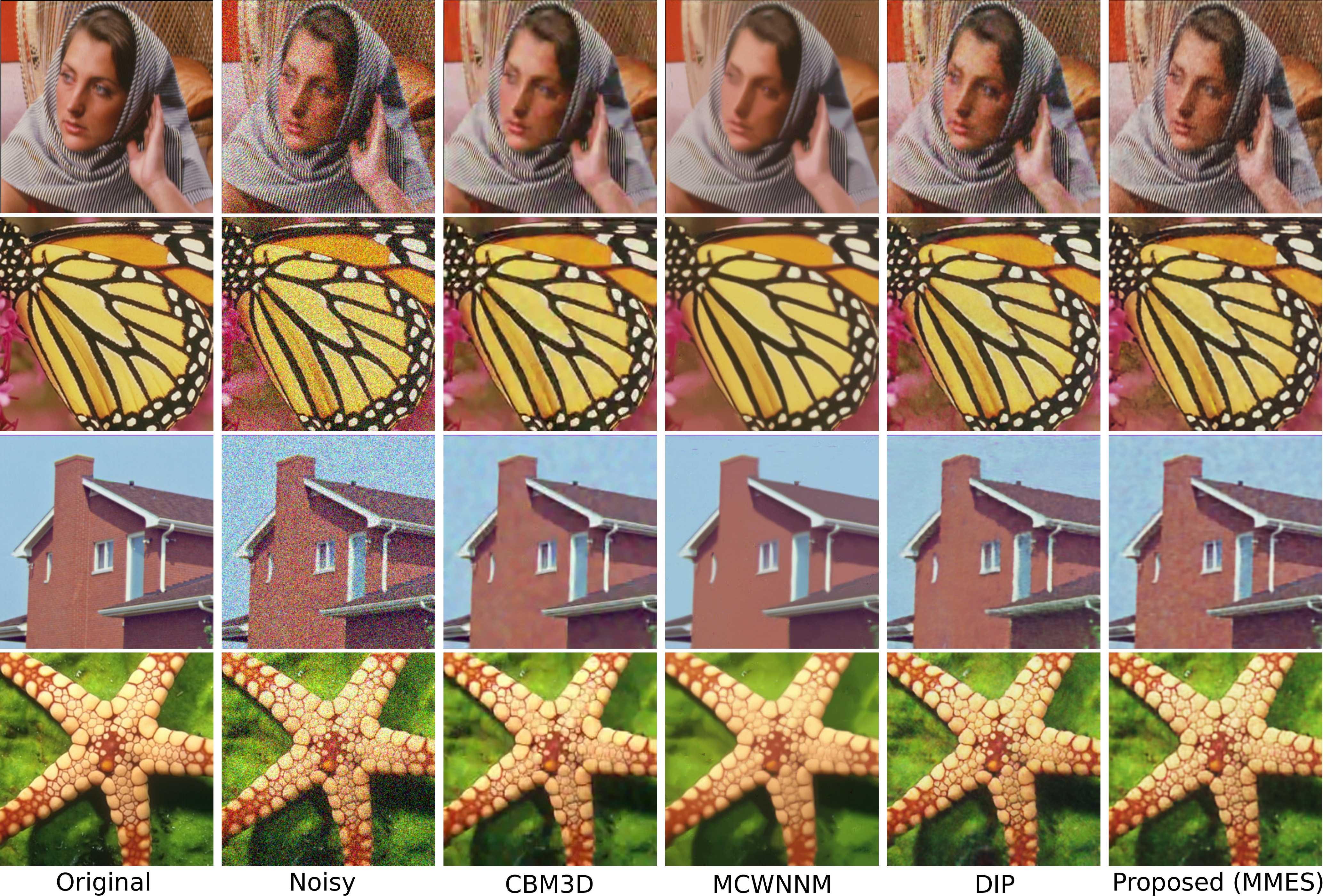}
  \caption{Denoising results in case of that the noise standard deviation is 40.}\label{fig:denoise}
\end{figure*}

\subsubsection{Color image denoising}\label{sec:denoise}
In this section, we compare performance of the proposed method with several selected unsupervised image denoising methods: CBM3D\footnote{For CBM3D, MATLAB software was downloaded from \url{http://www.cs.tut.fi/~foi/GCF-BM3D}.} \cite{dabov2007color}, MCWNNM\footnote{For MCWNNM, MATLAB software was downloaded from \url{https://github.com/csjunxu/MCWNNM_ICCV2017}.} \cite{xu2017multi}, and DIP \cite{ulyanov2018deep}.
We synthetically generated noisy images by additive Gaussian noise with various standard deviation from \{20, 30, 40, 50, 60, 70,\}.

In this experiments, both DIP and MMES were conducted with the best number of iterations from \{100, 200, 300, ..., 9900\} in the same early stopping strategy.
For all images in MMES, the architecture of auto-encoder consists of three hidden layers with sizes of $[8\tau^2, r, 8\tau^2]$, and parameters were set as $\tau=6$, $r=36$, and $\sigma=0.05$.
The trade-off parameter $\lambda$ was controlled to keep the low auto-encoder loss $\mathcal{L}_{\text{DAE}}$ and to minimize smoothly the reconstruction loss $\mathcal{L}_{\text{rec}}$.

Fig.~\ref{fig:psnr_bar}(c) shows the seven test images and averages of PSNR and SSIM in various noise levels.
According to this quantitative evaluation, CBM3D was the best in PSNR and SSIM, and DIP was slightly worse than other methods.
MCWNNM and MMES were competitive.

Fig.~\ref{fig:denoise} shows selected images reconstructed by four denoising methods.
MCWNNM reconstructed natural smooth images, but it tends to over-remove the signal components.
By contrast, DIP and MMES tends to leave noise components.
CBM3D provided the good balance to removing noise while keeping signal components.
Note that these denoising results will not show the essential superiority of the methods/models because of the possibility of non-optimal hyper parameter settings.

From the results, DIP and MMES were still similar in denoising problem.
Both methods have the same limitation about the difficulty of early stopping strategy, which is essentially equivalent to the difficulty of hyper-parameter tuning in denoising problem such as noise estimation and rank estimation.

\section{Interpretation of MMES toward explaining DIP}\label{sec:interpretation}
It is well known that there is no mathematical definition of interpretability in machine learning and there is no one unique definition of interpretation. We understand the interpretability as a degree to which a human can consistently predict the model's results or performance. The higher the interpretability of a deep learning model, the easier it is for someone to comprehend why certain performance or predictions or expected output can be achieved. We think that a model is better interpretable than another model if its performance or behaviors are easier for a human to comprehend than performance of the other models.

\subsection{From a perspective of dimensionality reduction/manifold learning}
The manifold learning and associated auto-encoder (AE) can be viewed as the generalized non-linear version of principal component analysis (PCA). In fact, manifold learning solves the key problem of dimensionality reduction very efficiently. In other words, manifold learning (modeling) is an approach to non-linear dimensionality reduction. Manifold modeling for this task is based on the idea that the dimensionality of many data sets is only artificially high. Although the patches of images (data points) consist of hundreds/thousands pixels, they may be represented as a function of only a few or quite limited number underlying parameters. That is, the patches are actually samples from a low-dimensional manifold that is embedded in a high-dimensional space. Manifold learning algorithms attempt to uncover these parameters in order to find a low dimensional representation of the images.

In our MMES approach to solve the problem we applied original embedding via multi-way delay embedding transform (MDT or Hankelization). Our algorithm is based on the optimization of cost function and it works towards extracting the low-dimensional manifold that is used to describe the high-dimensional data. The manifold is described mathematically by Eq.~\eqref{eq:def_manifold} and cost function is formulated by Eq.~\eqref{eq:cost}.

\subsection{Regarding our attempt to interpret "noise impedance in DIP" via MMES}
As mentioned at introduction, \cite{ulyanov2018deep} reported an important phenomenon of noise impedance of ConvNet structures, and our experiments demonstrated that the MMES has noise impedance too shown in Fig.~\ref{fig:impedance}.
Here, we provide a prospect for explaining the noise impedance in DIP through the MMES.

Let us consider the sparse-land model \cite{elad2006image_cvpr,elad2006image_tip}, i.e. noise-free images are distributed along low-dimensional manifolds in the high-dimensional Euclidean space and images perturbed by noises thicken the manifolds (make the dimension of the manifolds higher). Under this model, the distribution of images can be assumed to be higher along the low-dimensional noise-free image manifolds. When we assume that the image patches are sampled from low-dimensional manifold like sparse-land model, it is difficult to put noisy patches on the low-dimensional manifold. Let us consider to fit the network for noisy images. In such case the fastest way for decreasing squared error (loss function) is to learn "similar patches" which often appear in a large set of image-patches. Note that finding similar image-patches for denoising is well-known problem solved, e.g., by BM3D algorithm, which find similar image patches by template matching. In contrast, our auto-encoder automatically maps similar-patches into close points on the low-dimensional manifold. When similar-patches have some noise, the low-dimensional representation tries to keep the common components of similar patches, while reducing the noise components. This has been proved by \cite{alain2014regularized} so that a (denoising) auto-encoder maps input image patches toward higher density portions in the image space. In other words, a (denoising) auto-encoder has kind of a force to reconstruct the low-dimensional patch manifold, and this is our rough explanation of noise impedance phenomenon. Although the proposed MMES and DIP are not completely equivalent, we see many analogies and similarities and we believe that our MMES model and associated learning algorithm give some new insight for DIP.

\section{Discussions and Conclusions}

A beautiful manifold representation of complicated signals in embedded space has been originally discovered in a study of dynamical system analysis (\ie chaos analysis) for time-series signals \cite{packard1980geometry}.
After this, many signal processing and computer vision applications have been studied but most methods have considered only linear approximation because of the difficulty of non-linear modeling \cite{van1991subspace,szummer1996temporal,li1997parameter,ding2007rank,markovsky2008structured}.
However nowadays, the study of non-linear/manifold modeling has been well progressed with deep learning, and it was successfully applied in this study.
Interestingly, we could apply this non-linear system analysis not only for time-series signals but also natural color images and tensors (this is an extension from delay-embedding to multi-way shift-embedding).
The best of our knowledge, this is the first study to apply Hankelization with AE into general tensor data reconstruction.

MMES is a novel and simple image reconstruction model based on the low-dimensional patch-manifold prior which has many connections to ConvNet.
We believe it helps us to understand how work ConvNet/DIP through MMES, and support to use DIP for various applications like tensor/image reconstruction or enhancement \cite{gong2018pet,yokota2019dynamic,van2018compressed,Gandelsman_2019_CVPR}.

Finally, we established bridges between quite different research areas such as the dynamical system analysis, the deep learning, and the tensor modeling.
The proposed method is just a prototype and can be further improved by incorporating other methods such as regularizations, multi-scale extensions, and adversarial training.

\section*{Acknowledgment}
This work was partially supported by JST ACT-I Grant Number JPMJPR18UU, THE HORI SCIENCES AND ARTS FOUNDATION, and the MES RF grant 14.756.31.0001.

\bibliographystyle{ieee}

\end{document}